\documentclass[conference]{IEEEtran}
\IEEEoverridecommandlockouts
\usepackage{cite}
\usepackage{amsmath,amssymb,amsfonts}
\usepackage{algorithmic}
\usepackage{graphicx}
\usepackage{textcomp}
\usepackage{xcolor}
\def\BibTeX{{\rm B\kern-.05em{\sc i\kern-.025em b}\kern-.08em
    T\kern-.1667em\lower.7ex\hbox{E}\kern-.125emX}}

\usepackage[]{footmisc}

\usepackage{cite}
\usepackage{amsmath,amssymb,amsfonts}
\usepackage{algorithmic}
\usepackage{graphicx}
\usepackage{textcomp}
\usepackage{hyperref}
\usepackage{multirow}
\usepackage{array}

\newtheorem{definition}{Definition}[section]

\usepackage{bm}
\makeatletter
\AtBeginDocument{\DeclareMathVersion{bold}
\SetSymbolFont{operators}{bold}{T1}{times}{b}{n}
\SetMathAlphabet{\mathrm}{bold}{T1}{times}{b}{n}
\SetMathAlphabet{\mathit}{bold}{T1}{times}{b}{it}
\SetMathAlphabet{\mathbf}{bold}{T1}{times}{b}{n}
\SetMathAlphabet{\mathtt}{bold}{OT1}{pcr}{b}{n}
\SetSymbolFont{symbols}{bold}{OMS}{cmsy}{b}{n}
\renewcommand\boldmath{\@nomath\boldmath\mathversion{bold}}}
\makeatother

\usepackage{fancyhdr}
\fancypagestyle{firstpage}
{
    \fancyhead[L]{}    
    \fancyhead[L]{\textit{This work has been published in IEEE Access with DOI 10.1109/ACCESS.2024.3435558} \textit{and can be accessed online at }\href{https://ieeexplore.ieee.org/document/10614179}{ieeexplore.ieee.org/document/10614179}.}
}

\begin{document}

% Clarification of being a preprint submitedto an IEEE journal
\pagestyle{fancy}
\fancyfoot{}
\fancyfoot[LE,RO]{\thepage}
% \fancyfoot{}
% \fancyfoot[LO,CE]{\textit{This work has been submitted to the IEEE for possible publication. Copyright may be transferred without notice, after which this version may no longer be accessible.}}

\title{Knowledge Transfer for Cross-Domain Reinforcement Learning: A Systematic Review
\thanks{This work was supported with the PhD scholarship granted by CONAHCYT to Sergio A. Serrano.\\
$^1$ Computer Science Department, Instituto Nacional de Astrofísica, Óptica y Electrónica, Luis Enrique Erro \#1, San Andrés Cholula, 72840, Puebla, México (e-mail: sserrano@inaoep.mx, carranza@inaoep.mx, esucar@inaoep.mx)\\
$^2$ Computer Science Department, University of Bristol, Bristol BS8 1TL, Bristol, BS8 1TH, South West, England\\
$^{\dag}$ Corresponding author.
}
}

\author{Sergio A. Serrano$^1$, Jose Martinez-Carranza$^{1,2}$, and L. Enrique Sucar$^{1,\dag}$}

\maketitle

\begin{abstract}
Reinforcement Learning (RL) provides a framework in which agents can be trained, via trial and error, to solve complex decision-making problems. Learning with little supervision causes RL methods to require large amounts of data, rendering them too expensive for many applications (\textit{e.g.} robotics). By reusing knowledge from a different task, knowledge transfer methods present an alternative to reduce the training time in RL. Given the severe data scarcity, due to their flexibility, there has been a growing interest in methods capable of transferring knowledge across different domains (\textit{i.e.} problems with different representations). However, identifying similarities and adapting knowledge across tasks from different domains requires matching their representations or finding domain-invariant features. These processes can be data-demanding, which poses the main challenge in cross-domain knowledge transfer: to select and transform knowledge in a data-efficient way, such that it accelerates learning in the target task, despite the presence of significant differences across problems (\textit{e.g.} robots with distinct morphologies). Thus, this review presents a unifying analysis of methods focused on transferring knowledge across different domains. Through a taxonomy based on a transfer-approach categorization and a characterization of works based on their data-assumption requirements, the contributions of this article are 1) a comprehensive and systematic revision of knowledge transfer methods for the cross-domain RL setting, 2) a categorization and characterization of such methods to provide an analysis based on relevant features such as their transfer approach and data requirements, and 3) a discussion on the main challenges regarding cross-domain knowledge transfer, as well as on ideas of future directions worth exploring to address these problems.
\end{abstract}

\begin{IEEEkeywords}
Reinforcement Learning, Transfer Learning, Imitation Learning, Cross Domain, Review, Survey.
\end{IEEEkeywords}

\thispagestyle{firstpage}
\section{Introduction}\label{sec:introduction}
Developing technology that benefits society at large is the primary goal of human innovation. To this end, there are multiple ways in which artificial decision-making agents can improve life quality and safety in the workplace. Whether it is exploring dangerous environments (\textit{e.g.} outer space \cite{gao2017review}, radioactive zones \cite{nagatani2013emergency}, deep sea \cite{li2023bioinspired}), performing repetitive, labor-intensive chores \cite{bogue2016growth}, or taking care of the elder \cite{tanioka2019nursing}, building decision-making agents has the potential of helping humans in all of these assignments. Computers and robots can increase efficiency, precision, and consistency (compared to human performance) because they do not get tired or bored from repetitive activities. Moreover, artificial agents can make decisions faster, even if complex calculations are required. However, methods that yield safe and robust behaviors must be developed to delegate any of these tasks to robots.

\begin{figure}[t]
    \centering
    \includegraphics[width=0.4\textwidth]{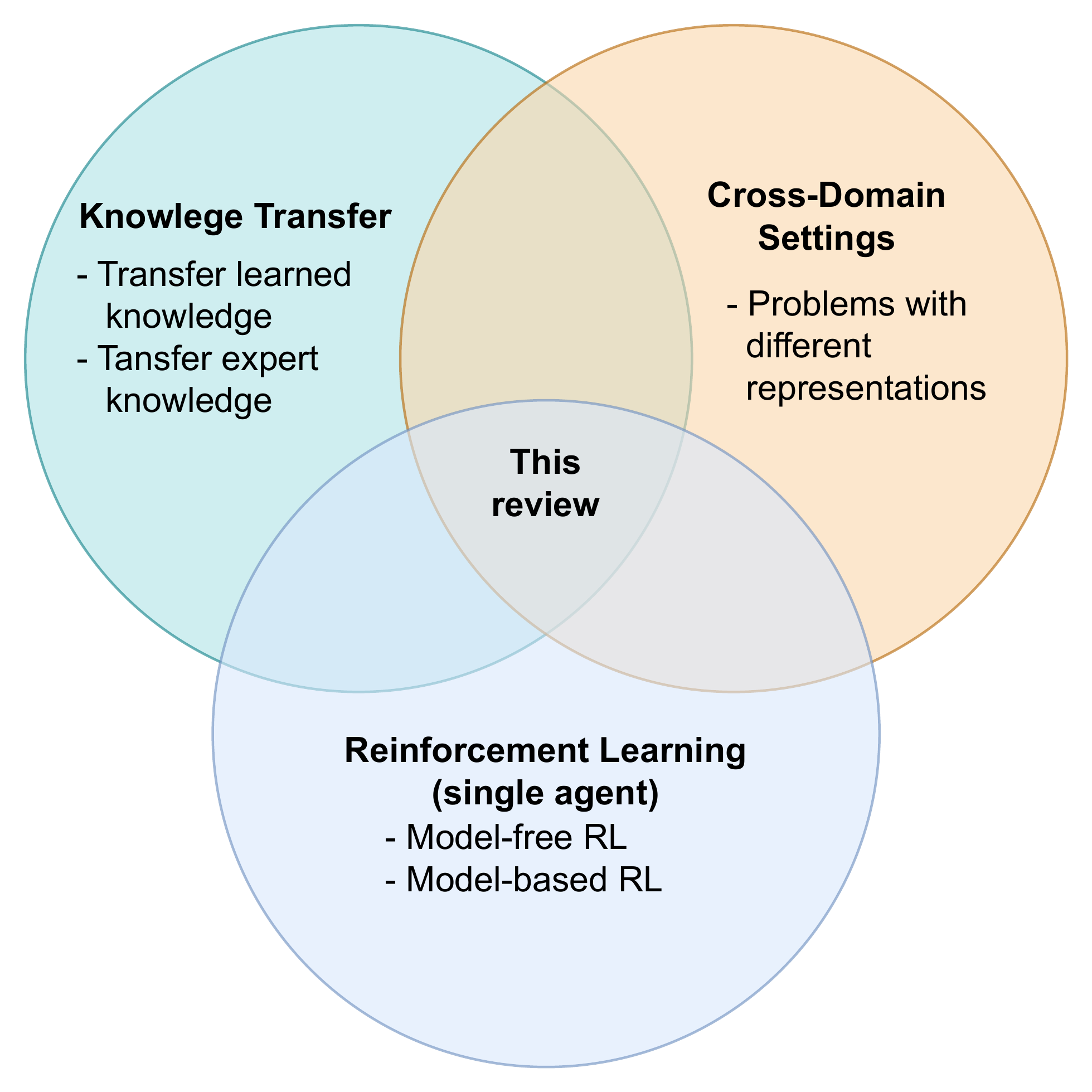}
    \caption{This review covers works that study transferring knowledge across single-agent reinforcement learning domains that have different state and/or action representations.}
    \label{fig:scope}
\end{figure}

By modeling how an agent perceives and causes changes in its environment and the dynamics that govern the immediate result of interactions, Reinforcement Learning (RL) presents a framework well-suited to describe sequential decision-making problems. Through exploring the environment and the weak supervision of a reward signal, RL agents can be trained to develop expert-like behaviors \cite{puterman2014markov}. On the other hand, despite how attractive learning with little supervision is, RL algorithms may require large amounts of data, which renders them incompatible with applications in which performing numerous interactions is unrealistic (\textit{e.g.} robotics due to the wear and tear effect \cite{kober2013reinforcement}).

To mitigate the limitations imposed by data costs, recent progress of simulators with realistic physics engines has extended the applicability of RL agents \cite{todorov2012mujoco,makoviychuk2021isaac}. In the case of problems in which a realistic simulator of the environment is not available, knowledge transfer methods present a viable alternative to decrease data costs by reusing knowledge from a related task \cite{taylor2009trlsurvey}. However, most knowledge transfer methods work under the assumption that if two tasks are similar enough, the knowledge acquired in one will be compatible with the other. 

The extent to which tasks are expected to be similar (or allowed to differ) depends on the knowledge transfer method used. A common assumption is that RL tasks have differences between the transition and/or reward dynamics, while the state and action spaces are shared \cite{konidaris2014hidden}. An example of this scenario would be a food-serving robot trained to use a tray to serve meals to diners. Given how liquids shift their weight distribution when moved, learning to serve food and beverages requires different behaviors. Similarly, if scorching plates are being served, it would be essential for the waiter robot to prioritize the diners' safety, even if dishes take longer to serve. Although differences in the meal fluidity (\textit{i.e.} transition dynamics) and additional safety restrictions (\textit{i.e.} reward dynamics) pose different tasks, knowledge transfer methods can exploit the commonalities (\textit{i.e.} the robot remains the same) to reuse expert behaviors across settings.

If a different robot needs to be trained, the knowledge transfer method must overcome the challenges presented by the mismatch between older and newer robots' actuators and sensors. What is more, if there are multiple robot models trained to serve food, a process to select from which robot knowledge should be transferred would be necessary. The \textit{selection} and \textit{adaptation} of knowledge across domains with significantly different representations are the core problems of cross-domain knowledge transfer. Whether it is combining \cite{ammar2015autonomous,qian2020intra,heng2022crossdomain,liu2023crptpro}, or filtering sources of knowledge \cite{talvitie2007experts,serrano2021inter,serrano2023similarity,zhang2024heterotrl}, the selection process is responsible for avoiding harming the learning of the recipient end (\textit{i.e.} negative transfer). On the other hand, adaptation must transform knowledge to become domain invariant while remaining relevant to the task.

Although the fundamental research questions remain untouched concerning same-domain knowledge transfer (\textit{i.e.} how to select and transfer knowledge), the representation mismatch forces us to rethink what we consider to be \textit{similar} and put it to the test in finding domain-invariant features and establishing a relation to the knowledge transfer process, so that training decision-making systems in complex environments becomes an affordable solution.

Therefore, we present a comprehensive review of transfer knowledge works for the cross-domain RL setting. The works covered in this document are organized in a feature-based taxonomy (see Table \ref{tab:taxonomy}) and compared by a set of dimensions related to the method requirements and problem resources (see Table \ref{tab:comparison}), to provide a characterization that helps the reader gain an insight of a method's approach, as well as how it compares to others.

\subsection{Paper Overview}
The contributions of our review are the following:
\begin{enumerate}
    \item A systematic revision of methods concerned with transferring different forms of knowledge across different domains in the reinforcement learning setting (see Fig. \ref{fig:scope}). For the sake of completeness, this review covers works that have already been analyzed in previous surveys \cite{bone2008trlsurvey}, \cite{taylor2009trlsurvey}, \cite{lazaric2012trlsurvey}, as well as the most recent developments.

    \item A feature-based taxonomy (see Table \ref{tab:taxonomy}) that classifies works based on their problem setting (\textit{e.g.} expert knowledge or acquired/learned knowledge), capabilities (\textit{e.g.} allows multiple or a single source of knowledge, requires state-only or state-action demonstrations) and transfer approach (\textit{e.g.} combines or filters sources of knowledge, transfers policies, parameters).

    \item A multidimensional comparison of the reviewed methods. To facilitate the selection of a transfer method that best suits the reader's needs, the comparative analysis (see Table \ref{tab:comparison}) emphasizes the requirements different methods have to work correctly, ranging from critical conditions (\textit{e.g.} knowledge quality) to more soft requisites (\textit{e.g.} inter-domain similarity assumptions).

    \item A discussion on the main open questions regarding knowledge transfer in cross-domain reinforcement learning and possible directions to follow in the short and long-term future. 
\end{enumerate}

The remainder of this article is structured as follows: Section \ref{sec:background} presents the basic concepts related to transfer learning, RL and a summary of other surveys that revise works in areas related to cross-domain knowledge transfer. Section \ref{sec:cdkt} introduces the definition of cross-domain knowledge transfer for RL that delimits the scope of this review and describes the taxonomy and dimensions of comparison, which we use for analysis and categorization purposes. The description of all the reviewed works is detailed in Sections \ref{sec:il} (Imitation Learning) and \ref{sec:tl} (Transfer Learning). Section \ref{sec:discussion} provides a discussion of the main open questions and future directions in cross-domain knowledge transfer for RL. Lastly, in Section \ref{sec:final-remarks}, the final remarks are presented.

\section{Background}\label{sec:background}
\subsection{Reinforcement Learning}\label{sec:background-rl}
\begin{figure}[t]
    \centering
    \includegraphics[width=0.44\textwidth]{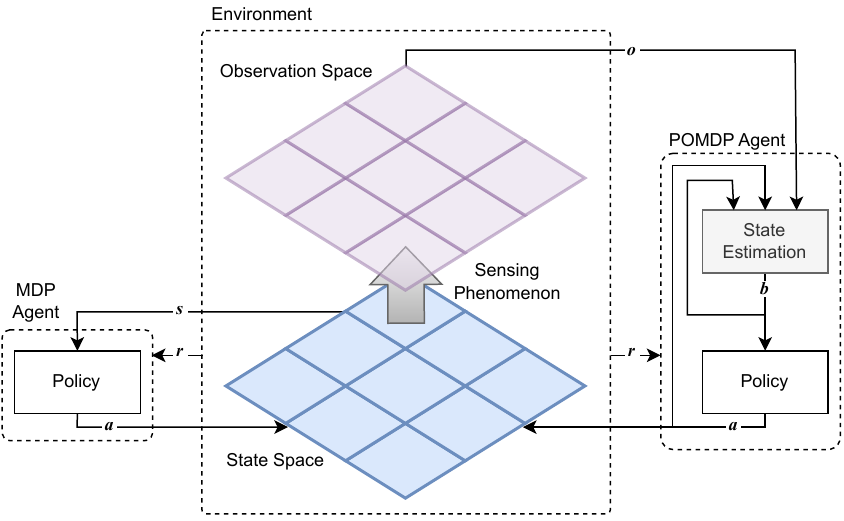}
    \caption{The difference between an MDP and POMDP setting is that the MDP agent has access to the system's state, whereas the POMDP agent perceives observations that depend on the system's state but do not fully describe it. The POMDP agent must learn a policy based on a constantly updated state estimation.}
    \label{fig:mdp-pomdp}
\end{figure}

Reinforcement Learning (RL) is the subarea of machine learning concerned with solving sequential decision-making problems in a semi-supervised setting. In RL, sequential decision-making problems are modeled as Markov Decision Processes (MDP), which describe the interactions between an agent and the environment/system (see Fig. \ref{fig:mdp-pomdp}) \cite{sutton2018reinforcement}. An MDP is a tuple $\langle S,A,\Phi,R,\gamma \rangle$ containing a state space $s\in S$, an action space $a \in A$, a transition function $\Phi : S \times A \times S \rightarrow [0,1]$ that describes the state transition probability given the executed action, a reward function $R : S \times A \rightarrow \mathbb{R}$ that returns a scalar (representing an immediate reward signal) and a discount factor $\gamma \in [0,1)$ that weights the present value of future rewards \cite{puterman2014markov}. Considering that an MDP policy $\pi : S \rightarrow A$ yields an action for every possible state, the RL objective is to find an optimal policy $\pi^\ast$ (Eq. \ref{eq:opt-policy}) that maximizes the state-value function, which is recursively defined by Bellman's equation (Eq. \ref{eq:bellman}).

\begin{equation}\label{eq:bellman}
    V^{\pi}(s) = max_a \bigg\{ R(s,a) + \gamma \sum_{s' \in S} \Phi(s' \mid s,a) V^{\pi}(s') \bigg\}
\end{equation}

\begin{equation}\label{eq:opt-policy}
    \pi^{\ast}(s) = argmax_a \bigg\{ R(s,a) + \gamma \sum_{s' \in S} \Phi(s' \mid s,a) V^{\pi^\ast}(s') \bigg\}
\end{equation}

The RL problem is mainly addressed by model-free and model-based reinforcement learning. Model-free RL includes methods that use state transitions and rewards to directly update a policy \cite{watkins1992q,sutton2018reinforcement}. In contrast, model-based methods learn an approximation of the reward or transition model as a step towards learning a value or policy function \cite{moerland2023model}.

Moreover, MDPs are a particular case of the Partially Observable MDPs, defined as a tuple $\langle S,A,\Phi,R,\gamma,O,\Omega,B_0 \rangle$, in which the agent does not have to the system's state. Instead, given the set of observations the agent can perceive $o \in O$, an observation function $\Omega : S \times A \times O \rightarrow [0,1]$ that models the probability of perceiving an observation after executing an action from a certain state, and an initial belief state distribution $B_0 : S \rightarrow [0,1]$ \cite{kaelbling1998planning,serrano2021knowledge}, the agent must learn an optimal policy based on an estimation of the system's state \cite{pineau2003point,spaan2005perseus,kurniawati2008sarsop,smith2012heuristic,shani2013survey} (see Fig. \ref{fig:mdp-pomdp}). Additionally, Table \ref{tab:variables} summarizes the variables frequently referenced in the description of the reviewed methods.

\begin{table}[t]
\caption{List of mathematical variables used for the description of RL systems throughout the reviewed works.}
\label{tab:variables}
\centering
\begin{tabular}{|cl|}
\hline
\multicolumn{1}{|l}{} & \multicolumn{1}{c|}{\textbf{Spaces}} \\
$S$ & State space \\
$A$ & Action space \\
$O$ & Observation space \\
$Z$ & Latent space \\
 & \multicolumn{1}{c|}{\textbf{Functions}} \\
$\Phi$ & State transition function \\
$R$ & Reward function \\
$\Omega$ & Observation function \\
$B_0$ & Initial state belief probability function \\
$V$ & State value function \\
$Q$ & Action value function \\
$\pi$ & Policy \\
 & \multicolumn{1}{c|}{\textbf{Data}} \\
$\langle s_i \rangle$ & Data set of states \\
$\langle s_i,a_i \rangle$ & Data set of state-action pairs \\
$\langle s_i,a_i,s_{i+1},r_i \rangle$ & Data set of interactions with the environment \\ \hline
\end{tabular}
\end{table}

% \Figure[t!]()[width=0.48\textwidth]{figures/mdp.png}{\textbf{Pomdps are cool.}\label{fig:mdp-pomdp}}

\subsection{Knowledge Transfer}\label{sec:background-tl}
\begin{figure}
    \centering
    \includegraphics[width=0.44\textwidth]{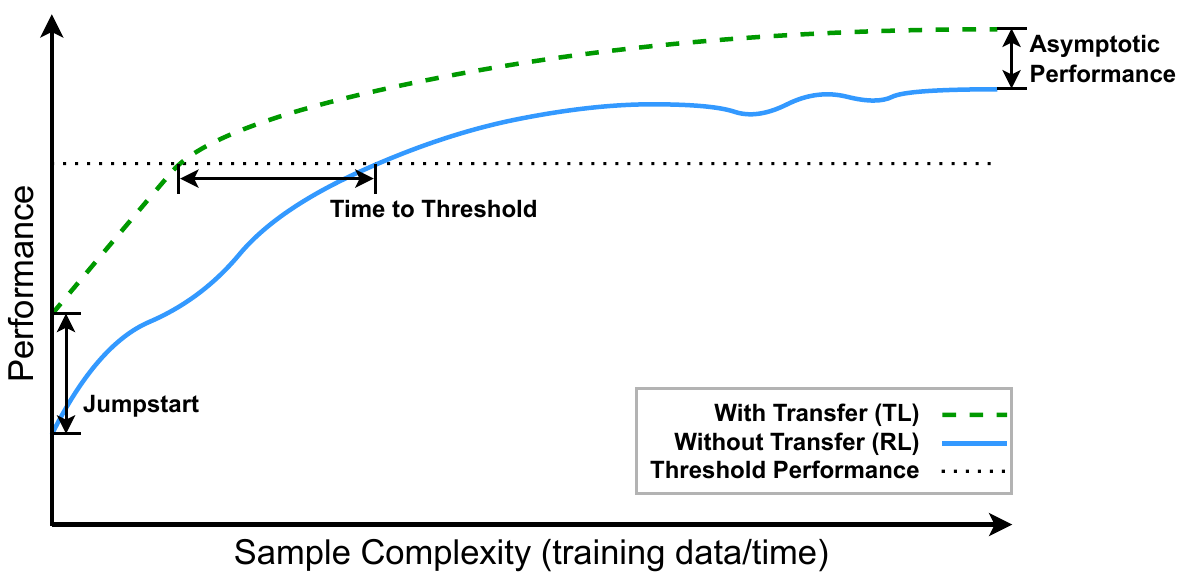}
    \caption{Transfer learning metrics evaluate the effect of transferring knowledge by measuring the performance difference, whether it is the initial performance (\textit{jumpstart}), final performance (\textit{asymptotic}), or the time it took to achieve certain performance (\textit{time to threshold}).}
    \label{fig:tl}
\end{figure}

Transfer Learning (TL) is concerned with studying methods that generalize knowledge previously acquired in a different problem (\textit{i.e., source task}) by transferring it to a new problem (\textit{i.e.} target task) to learn faster and/or better \cite{serrano2023similarity}. The performance of TL methods is evaluated by comparing the performance difference between reusing knowledge and learning from scratch. In \cite{taylor2009trlsurvey}, a collection of metrics is proposed to evaluate different aspects of TL approaches in the context of RL (see Fig. \ref{fig:tl}):

\begin{itemize}
    \item \textbf{Jumpstart}: The difference in initial performance.
    \item \textbf{Asymptotic Performance}: Describes how transferring knowledge affects the agent's final performance.
    \item \textbf{Total Reward}: The difference in total reward accumulated throughout the training process, \textit{i.e.} $R_{TL} - R_{RL}$.
    \item \textbf{Transfer Ratio}: the ratio of the \textbf{total reward metric} and total reward accumulated by the RL agent, \textit{i.e.} $\frac{R_{TL} - R_{RL}}{R_{RL}}$. In contrast to the \textbf{total reward metric}, the \textbf{transfer ratio} can be easily interpreted (as it describes a relative performance) without knowledge of the reward function.
    \item \textbf{Time to Threshold}: the training time or data (\textit{i.e.} sample complexity) required to achieve certain performance.
\end{itemize}

Methods that transfer knowledge are not limited to knowledge acquired by other learning algorithms but can also exploit knowledge provided by an expert or an oracle. For instance, in inverse reinforcement learning (IRL) \cite{ng2000algorithms,zhifei2012survey,arora2021survey} and imitation learning (IL) \cite{hussein2017imitation,osa2018algorithmic}, the objective is to learn to solve a task from demonstrations provided by an expert on such task. Similar to transfer learning methods, in both settings, the learning algorithm must adapt the expert demonstrations (source knowledge) into a format that allows solving the task at hand, whether it is learning a reward function under which the demonstrations display an optimal behavior (\textit{i.e.} IRL), or directly learning a policy that mimics the behavior that generated the demonstrations (\textit{i.e.} IL).

\subsection{Related Surveys}\label{sec:background-other-surveys}
The works covered in this review are categorized and compared with respect to a set of dimensions that we consider relevant for developing future knowledge transfer methods for cross-domain RL. However, other surveys overlap (to different degrees) with this article. For instance, \cite{bone2008trlsurvey}, \cite{taylor2009trlsurvey} and \cite{lazaric2012trlsurvey} review transfer learning works in a more general setting, which includes methods for both the same domain and cross domain, whereas \cite{niu2024cdtlsurvey} focus on policy transfer methods for the cross-domain setting. Similarly, \cite{glatt2016tdrlsurvey} and \cite{zhu2023tdrlsurvey} analyze transfer learning methods in the context of deep learning architectures.

Regarding multi-task RL, in \cite{vithayathil2020survey}, a review of methods that address the multi-task RL setting with deep learning is presented. Similarly, \cite{parisi2019continualsurvey} and \cite{khetarpal2022continualsurvey} cover methods for the lifelong/continual setting (which can be seen as an online multi-task problem). Other surveys that overlap with our review are revisions of Multi-agent RL \cite{da2019marlsurvey}, Hierarchical RL \cite{pateria2021hierarchicalsurvey}, Curriculum RL \cite{narvekar2020curriculumsurvey}, Simulation to Real Transfer RL \cite{salvato2021sim2realsurvey}. Moreover, in \cite{muller2021benchmarksurvey}, an analysis of the current state of benchmarks and evaluation methodologies for transfer learning is presented. 

From the surveys previously listed, \cite{niu2024cdtlsurvey} is the most similar to ours. However, they differ in two important aspects:
\begin{enumerate}
    \item \textbf{Type of Knowledge Transferred}: While we cover methods that transfer knowledge in a wide variety of forms, \cite{niu2024cdtlsurvey} focus on the comparison of TL methods that transfer policies across embodied agents (\textit{i.e.} robots).
    
    \item \textbf{Review Perspective}: The analysis presented in \cite{niu2024cdtlsurvey} is centered around the different types of gaps that might exist between different robotic applications (\textit{e.g.} type of sensors, sensor positions, length of robot limbs, number of sensors and actuators). On the other hand, the present review analyzes knowledge transfer works based on the transfer approach and the requirements a method has (\textit{e.g.} knowledge quality, inter-domain similarity assumptions), which determines the scenarios where it can be used.
\end{enumerate}

\section{Cross-Domain Knowledge Transfer}\label{sec:cdkt}
In the present review, we consider Cross-Domain Knowledge Transfer in RL to be the process of transferring any form of knowledge between problems with different representations. We define a domain in terms of the state and action spaces (see Definition \ref{def:domain}). By adding a source of supervision or reinforcement (\textit{e.g.} a reward function or expert demonstrations), a task can be defined in a domain (see Definition \ref{def:task}).

\begin{definition}[Domain]\label{def:domain}
    Let $S$ be a state space that describes the changeable aspects of a system and $A$ an action space containing the actions that can change the system's state, described by $S$. Then, a \textit{Domain} is a tuple $D = \langle S,A \rangle$.
\end{definition}

\begin{definition}[Task]\label{def:task}
    Let $Re$ be a reinforcement signal that guides the optimization process of an RL agent, defined as a reward function or (in the case of imitation learning) a set of demonstrations to be imitated. Thus, given a domain $D$, a \textit{Task} is a tuple $T = \langle D, Re \rangle$.
\end{definition}

For two domains to be considered different, a mismatch must exist between the state space, action space, or both. The mismatch (see Definition \ref{def:mismatch}) is present if a pair of spaces have different sizes, or they represent different phenomena (\textit{e.g.} configurations of different robots \cite{raychaudhuri2021cross} or different board games \cite{kuhlmann2007graph}). Thus, cross-domain knowledge transfer occurs when knowledge is shared across a pair of tasks whose state-action spaces have a mismatch (see Definition \ref{def:cross-dom-kt}).

\begin{definition}[Space Mismatch]\label{def:mismatch}
    Let $X$ and $Y$ be two spaces. We define a \textit{Space Mismatch} between $X$ and $Y$ as any of the following cases:
    \begin{enumerate}
        \item \textbf{Different Size}: $|X| \neq |Y|$ if both spaces are discrete and finite, or $m \neq n$ such that $X \subseteq \mathbb{R}^m, \: Y \subseteq \mathbb{R}^n$ if both spaces are continuous.

        \item \textbf{Different Phenomena}: $X$ and $Y$ represent different phenomena.
    \end{enumerate}
\end{definition}

\begin{definition}[Cross-Domain Knowledge Transfer]\label{def:cross-dom-kt}
    Let $T_1,T_2$ be two \textit{tasks} and their respective \textit{domains} $D_1,D_2$. \textit{Cross-Domain Knowledge Transfer} consists of transferring information that was acquired (or provided) while task $T_1$ was being learned to the learning process of task $T_2$, given that there is a \textit{space mismatch} between the state spaces or action spaces of domains $D_1$ and $D_2$.
\end{definition}

As a consequence of Definition \ref{def:cross-dom-kt} of cross-domain knowledge transfer, transferring knowledge across tasks that differ solely with respect to the transition or reward dynamics is not considered an instance of the cross-domain setting. Additionally, it is possible for tasks with a shared observation space to be considered a cross-domain instance only if their state-action spaces are different. For instance, \cite{liu2023crptpro} learn policies over an image-based representation of robots with different morphology.

\begin{table*}[h]
    \caption{Taxonomy of cross-domain knowledge transfer methods reviewed in this article. Knowledge transfer works are classified based on where the source knowledge comes from optimal demonstrations provided by an expert (imitation learning) or knowledge acquired by a learning method (transfer learning). Imitation learning methods are also separated into those that learn from demonstrations containing states and the actions that caused the state transitions (state-action demonstrations) or the works that only need sequences of states (state demonstrations). Similarly, transfer learning methods are further grouped depending on whether the method transfers knowledge from a single source (single-source transfer) or multiple sources (multiple-source transfer). Finally, single-source approaches are classified according to the main form of transferred knowledge. In contrast, multiple-source approaches are grouped based on how the source of knowledge is transferred: combining them into a single model (source combination) or selecting the best one (source selection). For a detailed description of each level of categorization, see Section \ref{sec:taxonomy}.}
    \label{tab:taxonomy}
    \begin{tabular}{@{}l@{}}
        $\mbox{\textbf{Cross-Domain Knowledge Transfer }}
        \left\{
        \begin{tabular}{@{}l@{}}
          $\mbox{\textbf{Imitation Learning} }
          \left\{
          \begin{tabular}{@{~}l@{}}
            \textbf{State-Action Demonstrations} (Section \ref{sec:il-sa}) \\
            \cite{kim2020domain}, \cite{fickinger2021cross} \\[\bigskipamount]
            \textbf{State Demonstrations}  (Section \ref{sec:il-s})\\
            \cite{raychaudhuri2021cross}, \cite{franzmeyer2022learn}, \cite{zakka2022xirl}, \cite{salhotra2023learning}, \cite{li2023crossloco}, \cite{li2023ace} \\
          \end{tabular}
          \right.$
          
          \hspace{-\nulldelimiterspace} \\[\bigskipamount]
          
          $\mbox{\textbf{Transfer Learning}}
          \left\{
          \begin{tabular}{@{~}l@{}}
            \textbf{Single-Source Transfer}
            $\left\{
              \begin{tabular}{@{~}l@{}}
                \textbf{Demonstrations} (Section \ref{sec:tl-demos}) \\
                \cite{ammar2012sparsecoding}, \cite{ammar2015unsupervised}, \cite{shankar2022translating}, \cite{aktas2023correspondence}, \cite{watahiki2023leveraging} \\[\bigskipamount]
                \textbf{Policy} (Section \ref{sec:tl-policy})\\
                \cite{soni2006homomorphism}, \cite{cheng2018reusing}, \cite{joshi2018cross}, \cite{shoeleh2020skill}, \cite{zhang2021learning}, \cite{wang2022weakly}, \cite{yang2023learn}, \cite{gui2023cross}, \cite{chen2024mirage} \\[\bigskipamount]
                \textbf{Reward Shaping} (Section \ref{sec:tl-shape}) \\
                \cite{brys2015policy}, \cite{gupta2017learning}, \cite{hu2019skill}, \cite{hejna2020hierarchically} \\[\bigskipamount]
                \textbf{Parameters} (Section \ref{sec:tl-params}) \\
                \cite{taylor2005valuemethods}, \cite{taylor2005behavior}, \cite{taylor2007tvitmps}, \cite{devin2017learning}, \cite{chen2019learning}, \cite{zhang2021feature} \\[\bigskipamount]
                \textbf{Value Function} (Section \ref{sec:tl-value}) \\
                \cite{torrey2006skill}, \cite{taylor2007cross}, \cite{taylor2007transfer}, \cite{taylor2007representation}, \cite{banerjee2007general}, \cite{kuhlmann2007graph}, \cite{torrey2008relational}, \cite{taylor2008autonomous} \\[\bigskipamount]
                \textbf{Bias} (Section \ref{sec:tl-bias}) \\
                \cite{torrey2005advice}, \cite{ammar2012commonTL}, \cite{wan2020mutual} \\[\bigskipamount]
                \textbf{Data} (Section \ref{sec:tl-data}) \\
                \cite{taylor2008instances}, \cite{cao2022learning}
              \end{tabular}
            \right.$
            \\[\bigskipamount]
            \textbf{Multiple-Source Transfer}
            $\left\{
              \begin{tabular}{@{~}l@{}}
                \textbf{Source Combination} (Section \ref{sec:tl-combine}) \\
                \cite{ammar2015autonomous}, \cite{qian2020intra}, \cite{heng2022crossdomain}, \cite{liu2023crptpro} \\[\bigskipamount]
                \textbf{Source Selection} (Section \ref{sec:tl-select}) \\
                \cite{talvitie2007experts}, \cite{serrano2021inter}, \cite{serrano2023similarity}, \cite{zhang2024heterotrl} \\
              \end{tabular}
            \right.$
            
          \end{tabular}
          \right.$
          
          \hspace{-\nulldelimiterspace}
        \end{tabular}
        \right.$
    \end{tabular}
\end{table*}

\subsection{Taxonomy}\label{sec:taxonomy}
Table \ref{tab:taxonomy} presents a classification of the knowledge transfer works covered in this article. The taxonomy works as a map, or index, of the reviewed works and seeks to assist the reader in quickly finding methods through three levels of categorization:

\begin{enumerate}
    \item \textbf{Origin of Knowledge}: First, knowledge transfer methods are classified based on where the source knowledge comes from, whether it is knowledge provided by an expert entity (\textit{i.e.} imitation learning) or knowledge was previously acquired/learned by a learning algorithm (\textit{i.e.} transfer learning).

    \item \textbf{Type of Source Knowledge}: The second level of categorization groups methods according to the type of source knowledge that is available to the method. In the case of imitation learning, having state-only or state-action demonstrations represents a significant gap in information. In contrast, transfer learning methods that transfer from multiple sources of knowledge face challenges different from the single-source approaches.

    \item \textbf{Transfer Approach}: In the third level, works are classified based on the approach employed to transfer knowledge. While single-source methods are categorized by the type of knowledge they transfer, multiple-source works are clustered based on how they manage the multiple sources, whether it is by combining them or selecting the \textit{best} source of knowledge to transfer from.
\end{enumerate}

\subsection{Approaches Overview}
Although the methods covered in this review were devised to address the restraints imposed by the data cost of standard RL algorithms, the variety in approaches adopted by each author presents different strengths and limitations. In order to aid in finding a method that suits a particular RL problem, the main (dis)advantages of three categories of the reviewed approaches are summarized below:

\begin{itemize}
    \item \textbf{Imitation Learning} (Section \ref{sec:il}): Imitation learning provides a framework to efficiently learn control schemes for RL tasks from expert demonstrations when the reward function is insufficient (\textit{e.g.} learning with sparse rewards \cite{vecerik2017leveraging}) or not available. However, the performance of IL methods strongly depends on the variety of the training data \cite{hua2021learning}. Moreover, a common assumption IL works make is that the demonstrations are sampled from a first-person perspective, which, unless there is a method in place to map data from different perspectives to the agent's point of view, can hinder adopting such an approach.

    \item \textbf{Transfer Learning with a Single Source of Knowledge} (Section \ref{sec:tl-single}): In contrast to learning from expert demonstrations, transfer learning approaches are an excellent fit for reusing knowledge acquired from tasks that are related to the one being learned. If enough domain-specific expertise is available to identify the \textit{relatedness} across tasks, methods that transfer from a single source of knowledge can accelerate learning \cite{taylor2009trlsurvey}. However, given that these methods do not have a way to assess the inter-task similarity, their efficacy may suffer if tasks are not related in a way that the method was developed to exploit (\textit{e.g.} similar reward/transition dynamics \cite{taylor2008instances}).

    \item \textbf{Transfer Learning with Multiple Sources of Knowledge} (Section \ref{sec:tl-multi}): When there are multiple sources of knowledge available to transfer knowledge from, but there is also uncertainty about what should be transferred, the methods reviewed in Section \ref{sec:tl-multi} present strategies to automate the reuse of multiple sources of knowledge, either by combining them, or selecting the best one for the task at hand. Nevertheless, these methods are still prone to harm the performance of the learning agent due to task interference (\textit{i.e.} degradation of performance after combining conflicting pieces of information) \cite{kanakis2020reparameterizing}, or because of the negative transfer as a consequence of selecting an unrelated source task \cite{taylor2009trlsurvey}.
\end{itemize}

\subsection{Dimensions of Comparison}\label{sec:cdkt-dim}
In order to compare cross-domain knowledge transfer works, a set of attributes was selected to characterize each method in terms of its knowledge requirements and knowledge transfer approach. The dimensions selected for the requirement analysis include Knowledge Quality, Inter-Domain Mapping, Similarity Assumptions, Allowed Learner and Source Selection. Based on the value in each dimension, it is possible to assess if a method is compatible with a particular problem setting. Moreover, the requirement dimensions are presented from most critical to least critical (from left to right) in Table \ref{tab:comparison} to facilitate readability.

Because the form of the transferred knowledge can be a crucial aspect to finding a solution for a particular application, the third-to-last dimension column in Table \ref{tab:comparison} describes the type of knowledge being transferred. Additionally, for complementary purposes, the last two columns describe the tasks in which the method was evaluated and the list of works (included in this review) the method experimentally compared to. The following sections describe in detail each dimension and the set of possible values they may hold.

\subsubsection{Knowledge Quality}
In order to successfully aid the target learner with auxiliary knowledge, data from the source and target domains must be informative enough to: 1) adapt the source knowledge so that it fits the receiving end (\textit{i.e.} target task), or 2) verify there exists inter-domain compatibility. We specify the knowledge quality required by a method as a $(req_{S};req_{T})$ tuple, in which the source ($req_{S}$) and target requirements ($req_{T}$) are separated by a semicolon. The different knowledge quality types include:
\begin{itemize}
    \item $\bm{ex}$ or $\bm{ex_S}$: Expert-generated demonstrations, represented as state-action trajectories $\langle ...,s_i,a_i,s_{i+1},a_{i+1},... \rangle$. If demonstrations include only state observations $\langle ...,s_i,s_{i+1},... \rangle$, then we use $\bm{ex_S}$.

    \item $\bm{i}$: Imperfect demonstrations. Data set of state-action transitions $\langle ...,s_i,a_i,s_{i+1},a_{i+1},... \rangle$, generated by a variety of unknown actors (which may include experts and random policies), whose quality is uncertain. We use the superscripted term $\bm{i^l}$ in case demonstrations are labeled with a quality score.

    \item $\bm{en_{\{\ast,l,r\}}}$: Availability of the environment for the agent to interact with is assumed. If data is sampled by a completely trained policy, a learning policy, or a random policy, we use $\bm{en_\ast}$, $\bm{en_l}$, $\bm{en_r}$, respectively. Additionally, if the environment is required to be differentiable (\textit{e.g.} to compute the gradient of a loss function), then the nabla symbol precedes the environment abbreviation $\bm{\nabla en}$.

    \item $\bm{s}$: Availability of a simulator of the environment, in which the state can be set, and state transitions and rewards can be queried.

    \item $\bm{\pi}$: Availability of a trained policy.

    \item $\bm{m}$: Availability of a model that approximates the transition, or reward, function.

    \item $\bm{or}$: Access to oracle-level parameters, \textit{i.e.} parameters that govern the transition or reward function of the environment the agent will interact with.

    \item $\bm{px}$: Set of data tuples $\langle s_{i},a_{i},s_{i+1},r_{i} \rangle$ sampled while learning a policy for a proxy task (\textit{i.e.} a task that is different from the target task).

\end{itemize}
In case the method requires data from the source and target domain to be paired in some way, we identify four levels of data pairing:
\begin{itemize}
    \item $\bm{o}$: Task-level overlapping. The method assumes that given a pair of collections of data sets sampled for multiple tasks in both domains, a subset of data sets in both collections were sampled while learning the same tasks. However, it is unknown which data sets match across collections.
    
    \item $\bm{t}$: Task-level pairing. A pair of data sets are assumed to be generated after sampling from two domains while solving (or learning to solve) the same task.

    \item $\bm{d}$: Demo-level pairing. In addition to having task-level pairing, trajectories corresponding to the same demonstration made in two domains are paired.

    \item $\bm{a}$ or $\bm{\hat{a}}$: Alignment. In addition to having demo-level pairing, the states, actions, or state-actions (in demonstrations from different domains) are paired to provide a perfect data alignment. If the pairings are also provided with a score (to reflect the quality/confidence of the pairing) or are partially aligned (\textit{i.e.} pairings are provided only for a subset of the demonstrations), then we use $\bm{\hat{a}}$ to represent the soft/partial alignment.
\end{itemize}
The data-pairing requirement is expressed as a superscript of the paired data, \textit{e.g.} a tuple $\bm{(ex,px^t;en_l,px^t)}$ describes that task-level paired data sets are needed, in addition to state-action expert demonstrations in the source domain, and interactions between the target domain environment and learning agent.

\subsubsection{Inter-Domain Mapping}
Type of inter-domain mapping used to establish correspondences between spaces of the source and target domain. A comma separates independent mappings of multiple spaces, while joint mappings are concatenated by the cross-product operator (\textit{i.e.} $\times$). Subscripts in space variables describe whether they belong to the source ($S$) or target domain ($T$). Mapped spaces include:
\begin{itemize}
    \item $\bm{S}$: State space.
    \item $\bm{A}$: Action space.
    \item $\bm{O}$: Observation space.
    \item $\bm{Z}$: Latent space, learned to capture domain-specific (\textit{i.e.} the domain it belongs to is specified as a subscript) or domain-invariant features (\textit{i.e.} does not have a subscript since both domains share it).
    \item $\bm{obj}$: Set of objects that partially determine the state of the system and that the agent manipulates.
    \item $\bm{none}$: The method does not use an inter-domain mapping.
\end{itemize}
In addition to specifying the direction in which information flows, the letter on top of an arrow describes the process employed to obtain the mapping:
\begin{itemize}
    \item $\bm{\underrightarrow{h}}$: The mapping was hand-coded and provided to the method.
    \item $\bm{\underrightarrow{l}}$: The mapping was learned by the method from data.
    \item $\bm{\underrightarrow{c}}$: A set of candidate mappings was provided, and the method decides which one to use.
\end{itemize}

\subsubsection{Similarity Assumptions}
Assumptions about the type of relation the source and target domains have. Similarity assumptions are a premise that must be met for the method to work. These assumptions are either explicitly made by the authors or are a condition imposed by the knowledge quality requirements (\textit{e.g.} paired data implies that a task in one domain has an equivalent in the other domain). We identify the following inter-domain similarity assumptions:
\begin{itemize}
    \item $\bm{ali}$: The source and target domain are aligned by a set of tasks that can be specified and solved in both domains. For instance, robot arms with different morphologies are aligned by tasks such as \textit{picking objects up}; however, that is not the case for a robot arm and a chess-playing agent.
    
    \item $\bm{lin}$: Task-specific policies can be expressed as a linear combination of a latent basis shared across domains.
    
    \item $\bm{same_{\{S,A,O\}}}$: Domains have the same state space ($\bm{same_S}$), action space ($\bm{same_A}$) or observation space ($\bm{same_O}$), while the other spaces are different.

    \item $\bm{share_S}$: Domains share a subset of state variables.
    
    \item $\bm{app}$: The method works on domains that belong to the same application (\textit{e.g.} turn-taking games, robotic manipulation of objects).
    
    \item $\bm{rel}$: The domain dynamics (\textit{i.e.} transition and reward functions) are conditioned on a shared latent structure (\textit{e.g.} the state transition dynamics of walking robots with different legs can be similar due to having an upper-torso).
    
    \item $\bm{none}$: No assumptions are made.
\end{itemize}

\subsubsection{Allowed Learner}
Type of learning architectures and algorithms the knowledge transfer method is compatible with:
\begin{itemize}
    \item $\bm{any}$: The knowledge transfer method interacts with the agent as a decision-making black box. Therefore, any architecture and algorithm can be used to model the learning agent.
    \item $\bm{nn}$: The agent is modeled as a neural network.
    \item $\bm{pg}$: The agent is learned via policy-gradient methods.
    \item $\bm{vf}$: The agent is modeled as a state-value function.
    \item $\bm{qf}$: The agent is modeled as an action-value function \cite{watkins1992q}.
    \item $\bm{ac}$: The agent is modeled with an actor-critic architecture (\textit{e.g.} SAC \cite{haarnoja2018soft,haarnoja2018softalgo}).
    \item $\bm{mb}$: A model-based learning algorithm is required.
    \item $\bm{ms}$: An architecture and algorithm that is specific to the knowledge transfer method must be employed to model the learning agent.
    \item $\bm{bc}$: Behavior cloning methods devised to replicate the policy that generated a data set are required.
    \item $\bm{opt}$: Options \cite{sutton1998intra,sutton1999between}, which are policies accompanied by a set of states from which they can be invoked and a set of termination conditions that determine when they must stop taking actions. Options are usually employed to model abstract actions in hierarchical decision-making systems.
\end{itemize}
Regarding knowledge transfer methods that allow for different learning architecture/algorithms in the source and target domain, a semi-colon will separate the source (left side) and target (right side) allowed learners. Otherwise, no parenthesis will be used. For instance, $\bm{(nn;pg,ac)}$ describes a knowledge transfer method that allows neural network architectures in the source domain, while policy gradient and actor-critic methods in the target domain side.

\subsubsection{Source Selection}
Process responsible for selecting the source(s) of knowledge from which knowledge will be transferred to the target domain agent:
\begin{itemize}
    \item $\bm{man}$: An oracle or expert manually selects the single source of knowledge the method will use.
    \item $\bm{auto}$: The method selects a single source of knowledge (at a time) from a set of candidates.
    \item $\bm{multi}$: The method combines multiple sources of knowledge from which knowledge will be simultaneously transferred.
\end{itemize}

\subsubsection{Knowledge Transferred}
A wide range of variables factor into an RL agent successfully learning to solve the task. Thus, in order to accelerate the learning process of the target agent, there are many ways in which auxiliary knowledge can help. We identify the following forms of transferred knowledge:
\begin{itemize}
    \item $\bm{demo}$ or $\bm{demo_S}$: Set of expert-generated demonstrations in the form of $(s_{i},a_{i},s_{i+1})$ tuples. If actions are not available in the demonstration (\textit{i.e.} state-only transitions $(s_{i},s_{i+1})$), then we use $\bm{demo_S}$.
    
    \item $\bm{shape}$: Reward shaping, which modifies (or replaces) the original reward function via an auxiliary reward term.
    
    \item $\bm{\pi}$: Policy adapted to make decisions in the target domain.
    
    \item $\bm{par}$: Parameters that characterize (some of) the agent's policy (\textit{e.g.} actor, critic) or model of the environment (\textit{e.g.} a neural network that approximates the transition function).
    
    \item $\bm{bias_{\{\pi,par\}}}$: Inductive bias that influences how the policy ($\bm{bias_{\pi}}$) or parameters of a particular model ($\bm{bias_{par}}$) evolve during the learning process of the target-domain agent.
    
    \item $\bm{data}$: Set of $(s_{i},a_{i},s_{i+1},r_{i})$ or $(s_{i},a_{i},s_{i+1})$ data tuples sampled by a policy whose level of expertise is unspecified. Thus, they can not be considered expert demonstrations.
    
    \item $\bm{ann}$: Oracle/expert-given annotations that describe the quality of a data set. For instance, the labels for a data set of imperfect demonstrations that specify how similar they are to an optimal behavior.
    
    \item $\bm{Q}$: Action values of multiple state-action pairs.
    
    \item $\bm{V}$: State values of multiple states.
\end{itemize}

\subsubsection{Tasks}
Benchmark tasks, popular robotic platforms, and variations that were used to evaluate knowledge transfer methods in RL include:
\begin{itemize}
    \item $\bm{mj}$: Mujoco-simulated control tasks in the Gym suite \cite{todorov2012mujoco,brockman2016openai}, which include (but are not limited to) ant, half cheetah, walker, hopper, swimmer, reacher, inverted pendulum.
    
    \item $\bm{dm}$: Deep mind control suite \cite{tassa2018deepmind} which includes control problems for visual-oriented agents.

    \item $\bm{sc}$: Soccer tasks such as keep away and break away \cite{stone2006keepaway}.

    \item $\bm{cc}$: Classic control problems which include mass-spring-damp system \cite{burchett2005parametric}, mountain car \cite{moore90efficientmemory-based}, cart pole \cite{chung1995nonlinear}.

    \item $\bm{rp}$: Robotic platforms such as spot\footnote{\href{https://bostondynamics.com/}{https://bostondynamics.com/}}, aliengo\footnote{\href{https://m.unitree.com/aliengo/}{https://m.unitree.com/aliengo/}}, sawyer\footnote{\href{https://www.rethinkrobotics.com/sawyer}{https://www.rethinkrobotics.com/sawyer}}, franka panda\footnote{\href{https://robodk.com/robot/Franka/Emika-Panda}{https://robodk.com/robot/Franka/Emika-Panda}}, some of which are set up in the \textit{Robosuite} framework \cite{zhu2020robosuite}.

    \item $\bm{pm}$: Point mass maze environments \cite{duan2016benchmarking,hejna2020hierarchically,fu2020d4rl} in which a point-body agent must learn to navigate (within a maze) to a final location.

    \item $\bm{xm}$: XMagical benchmark \cite{toyer2020magical,zakka2022xirl}, which includes agents with different body types whose goal is to move three blocks into a target zone.
    
    \item $\bm{bg}$: Board games \cite{banerjee2007general,kuhlmann2007graph} such as tic-tac-toe, connect-$\{3,4\}$, captureGo, othello, checkers, minichess.

    \item $\bm{bw}$: Blocks world \cite{slaney2001blocks}, where the goal is to move blocks to reach a goal configuration.

    \item $\bm{df}$: Manipulation of deformable objects \cite{matas2018sim}, such as cloth folding.
\end{itemize}

\subsubsection{Comparison to Other Works}
The rightmost column of Table \ref{tab:comparison} shows a list of other works (reviewed in this document) to which the method experimentally compared. If no comparison to other reviewed methods was made, then a non-applicable (NA) abbreviation is in place.

\begin{table*}[p]
\scriptsize
\caption{Comparison of cross-domain knowledge transfer works (see Section \ref{sec:cdkt-dim} for column descriptions).}
\label{tab:comparison}
\centering
\begin{tabular}{c|ccccccccc}
\hline
Work & \begin{tabular}[c]{@{}c@{}}Knowledge\\ Quality\end{tabular} & \begin{tabular}[c]{@{}c@{}}Inter-Domain\\ Mapping\end{tabular} & \begin{tabular}[c]{@{}c@{}}Similarity\\ Assumptions\end{tabular} & \begin{tabular}[c]{@{}c@{}}Allowed\\ Learners\end{tabular} & \begin{tabular}[c]{@{}c@{}}Source\\ Selection\end{tabular} & \begin{tabular}[c]{@{}c@{}}Knowledge\\ Transferred\end{tabular} & \begin{tabular}[c]{@{}c@{}}Tasks\end{tabular} & \begin{tabular}[c]{@{}c@{}}Comparison\end{tabular}\\
\hline
\multicolumn{9}{l}{Imitation Learning from States and Actions (Section \ref{sec:il-sa})} \\ \hline
\cite{kim2020domain} & $\bm{(px^t;px^t)}$ & $\bm{S_T \underleftrightarrow{l} S_S, A_S \underrightarrow{l} A_T}$  & $\bm{ali}$ & $\bm{(any;bc)}$ & $\bm{man}$ & $\bm{\pi,demo}$ & $\bm{cc,mj}$ & \cite{ammar2015unsupervised,gupta2017learning} \\
\cite{fickinger2021cross} & $\bm{(ex;en_l)}$ & $\bm{none}$ & $\bm{none}$ & $\bm{any}$ & $\bm{man}$ & $\bm{shape}$ & $\bm{pm,mj}$ & NA \\
\hline\multicolumn{9}{l}{Imitation Learning from States (Section \ref{sec:il-s})} \\ \hline
\cite{raychaudhuri2021cross} & $\bm{(px^t,ex_S;px^t,en_r)}$ & $\bm{S_S \underleftrightarrow{l} S_T}$ & $\bm{ali}$ & $\bm{bc}$ & $\bm{man}$ & $\bm{demo_S}$ & $\bm{mj}$ & \cite{gupta2017learning} \\
\cite{franzmeyer2022learn} & $\bm{(ex_S;en_l)}$ & $\bm{S_T \underrightarrow{l} S_S}$ & $\bm{ali}$ & $\bm{any}$ & $\bm{man}$ & $\bm{demo_S}$ & $\bm{mj,xm}$ & \cite{fickinger2021cross,zakka2022xirl} \\
\cite{zakka2022xirl} & $\bm{(ex_S;en_l)}$ & $\bm{S_S,S_T \underrightarrow{l} Z}$ & $\bm{ali}$ & $\bm{any}$ & $\bm{multi}$ & $\bm{shape}$ & $\bm{xm}$ & NA \\
\cite{salhotra2023learning} & $\bm{(ex_S;s)}$ & $\bm{none}$ & $\bm{same_S}$ & $\bm{bc}$ & $\bm{man}$ & $\bm{demo_S}$ & $\bm{rp,df}$ & NA \\
\cite{li2023crossloco} & $\bm{(ex_S;en_l)}$ & $\bm{S_S\underleftrightarrow{l} S_T}$ & $\bm{ali}$ & $\bm{any}$ & $\bm{man}$ & $\bm{shape}$ & $\bm{rp}$ & NA \\
\cite{li2023ace} & $\bm{(px;px)}$ & $\bm{S_S \underrightarrow{l} S_T}$ & $\bm{ali}$ & $\bm{bc}$ & $\bm{man}$ & $\bm{demo_S}$ & $\bm{rp}$ & NA \\
\hline\multicolumn{9}{l}{Transfer Learning from a Single Source - Demonstrations (Section \ref{sec:tl-demos})} \\ \hline
\cite{ammar2012sparsecoding} & $\bm{(px,\pi;px,s,en_l)}$ & $\bm{S_S^2 \times A_S \underleftrightarrow{l} S_T^2 \times A_T}$ & $\bm{none}$ & $\bm{any}$ & $\bm{man}$ & $\bm{demo}$ & $\bm{cc}$ & NA \\
\cite{ammar2015unsupervised} & $\bm{(en_{\ast};en_l)}$ & $\bm{S_T\underleftrightarrow{l} S_S}$ & $\bm{none}$ & $\bm{(any;pg)}$ & $\bm{man}$ & $\bm{demo_S}$ & $\bm{cc}$ & NA \\
\cite{shankar2022translating} & $\bm{(px^o,px^o)}$ & $\bm{Z_S\underrightarrow{l} Z_T}$ & $\bm{ali}$ & $\bm{any}$ & $\bm{man}$ & $\bm{demo}$ & $\bm{rp}$ & NA \\
\cite{aktas2023correspondence} & $\bm{(px^t;px^t)}$ & $\bm{S_S \underrightarrow{l} A_T}$ & $\bm{ali}$ & $\bm{ms}$ & $\bm{man}$ & $\bm{demo_S}$ & $\bm{rp}$ & NA \\
\cite{watahiki2023leveraging} & $\bm{(px^t,px^t)}$ & \begin{tabular}[c]{@{}c@{}} $\bm{\{S_S,S_T\} \underrightarrow{l} Z^X,}$ \\ $\bm{Z^A \underrightarrow{l} \{A_S,A_T\} }$ \end{tabular} & $\bm{ali}$ & $\bm{bc}$ & $\bm{man}$ & $\bm{\pi,demo}$ & $\bm{pm,rp}$ & \cite{kim2020domain,raychaudhuri2021cross} \\
\hline\multicolumn{9}{l}{Transfer Learning from a Single Source - Policy (Section \ref{sec:tl-policy})} \\ \hline
\cite{soni2006homomorphism} & $\bm{(\pi;en_l)}$ & $\bm{S_T \underrightarrow{c} S_S}$ & $\bm{ali}$ & $\bm{(opt;qf)}$ & $\bm{man}$ & $\bm{\pi}$ & $\bm{sc,bw}$ & NA \\
\cite{cheng2018reusing} & $\bm{(\pi;en_l)}$ & $\bm{S_T \underrightarrow{h} S_S}$ & $\bm{same_A}$ & $\bm{(qf;any)}$ & $\bm{man}$ & $\bm{\pi}$ & $\bm{sc}$ & \cite{taylor2007transfer} \\
\cite{joshi2018cross} & $\bm{(sim,\pi;en_r)}$ & $\bm{S_{S} \underleftrightarrow{l} S_{T},\:A_{S} \underleftrightarrow{h} A_{T}}$ & $\bm{rel}$ & $\bm{qf}$ & $\bm{man}$ & $\bm{\pi,bias_{\pi}}$ & $\bm{cc}$ & \cite{ammar2015unsupervised} \\
\cite{shoeleh2020skill} & $\bm{(en_l,\pi;en_r,en_l)}$ & $\bm{S_{T} \times A_{T} \underrightarrow{l} S_{S} \times A_{S}}$ & $\bm{none}$ & $\bm{opt,ac}$ & $\bm{man}$ & $\bm{\pi,Q}$ & $\bm{cc}$ & NA \\
\cite{zhang2021learning} & $\bm{(px,px)}$ & \begin{tabular}[c]{@{}c@{}} $\bm{S_T \underrightarrow{l} S_S}$, $\bm{S_T \times A_T \underrightarrow{l} A_S}$ \\ $\bm{S_S \times A_S \underrightarrow{l} A_T}$\end{tabular} & $\bm{rel}$ & $\bm{any}$ & $\bm{man}$ & $\bm{\pi}$ & $\bm{mj,dm}$ & NA \\
\cite{wang2022weakly} & $\bm{(px^{\hat{a}},px^{\hat{a}})}$ & \begin{tabular}[c]{@{}c@{}} $\bm{S_T \underrightarrow{l} S_S}$, $\bm{S_T \times A_T \underrightarrow{l} A_S}$ \\ $\bm{S_S \times A_S \underrightarrow{l} A_T}$\end{tabular} & $\bm{ali}$ & $\bm{any}$ & $\bm{man}$ & $\bm{\pi}$ & $\bm{mj,rp}$ & NA \\
\cite{yang2023learn} & $\bm{(px^t;px^t,en_l)}$ & $\bm{Z_{S} \underleftrightarrow{l} Z_{T}}$ & $\bm{ali,same_S}$ & $\bm{ac}$ & $\bm{man}$ & $\bm{\pi}$ & $\bm{rp}$ & NA \\
\cite{gui2023cross} & $\bm{(en_l,s;i)}$ & $\bm{S_T \underrightarrow{l} S_S,\: S_S \times A_S \underrightarrow{l} A_T}$ & $\bm{none}$ & $\bm{ms}$ & $\bm{man}$ & $\bm{\pi}$ & $\bm{mj}$ & NA \\
\cite{chen2024mirage} & $\bm{(s,or;s,or)}$ & $\bm{O_T \underrightarrow{h} O_S}$ & $\bm{same_A,app}$ & $\bm{any}$ & $\bm{man}$ & $\bm{\pi}$ & $\bm{rp}$ & NA \\
\hline\multicolumn{9}{l}{Transfer Learning from a Single Source - Reward Shaping (Section \ref{sec:tl-shape})} \\ \hline
\cite{brys2015policy} & $\bm{(\pi;en_l)}$ & $\bm{S_S \times A_S \underrightarrow{h} S_T \times A_T}$ & $\bm{none}$ & $\bm{any}$ & $\bm{man}$ & $\bm{shape}$ & $\bm{cc}$ & NA \\
\cite{gupta2017learning} & $\bm{(px^a,ex_S;px^a,en_l)}$ & $\bm{\{S_S,S_T\} \underrightarrow{l} Z}$ & $\bm{ali}$ & $\bm{any}$ & $\bm{man}$ & $\bm{shape}$ & $\bm{mj}$ & NA \\
\cite{hu2019skill} & $\bm{(px^a;px^a)}$ & $\bm{S_S \times A_S \underleftrightarrow{l} S_T \times A_T}$ & $\bm{ali,rel}$ & $\bm{any}$ & $\bm{man}$ & $\bm{shape}$ & $\bm{mj}$ & \cite{gupta2017learning} \\
\cite{hejna2020hierarchically} & $\bm{(\pi;en_l)}$ & $\bm{none}$ & $\bm{share_S}$ & $\bm{any}$ & $\bm{man}$ & $\bm{shape,bias_{\pi}}$ & $\bm{pm,mj}$ & NA \\
\hline\multicolumn{9}{l}{Transfer Learning from a Single Source - Parameters (Section \ref{sec:tl-params})} \\ \hline
\cite{taylor2005valuemethods} & $\bm{(\pi;en_l)}$ & $\bm{S_T \times A_T \underrightarrow{h} S_S \times A_S}$ & $\bm{ali}$ & $\bm{qf}$ & $\bm{man}$ & $\bm{par}$ & $\bm{sc}$ & NA \\
\cite{taylor2005behavior} & $\bm{(\pi;en_l)}$ & $\bm{S_T \times A_T \underrightarrow{h} S_S \times A_S}$ & $\bm{none}$ & $\bm{qf}$ & $\bm{man}$ & $\bm{par}$ & $\bm{sc}$ & NA \\
\cite{taylor2007tvitmps} & $\bm{(\pi;en_l)}$ & $\bm{S_T \times A_T \underrightarrow{h} S_S \times A_S}$ & $\bm{rel}$ & $\bm{nn}$ & $\bm{man}$ & $\bm{par}$ & $\bm{sc}$ & NA \\
\cite{devin2017learning} & $\bm{(\pi;\:)}$ & $\bm{none}$ & $\bm{share_S,ali}$ & $\bm{nn}$ & $\bm{man}$ & $\bm{par}$ & $\bm{mj}$ & NA \\
\cite{chen2019learning} & $\bm{(m,\pi;en_l)}$ & $\bm{none}$ & $\bm{rel}$ & $\bm{nn}$ & $\bm{man}$ & $\bm{par}$ & $\bm{mj}$ & \cite{wan2020mutual} \\
\cite{zhang2021feature} & $\bm{(\pi;en_l)}$ & $\bm{none}$ & $\bm{rel}$ & $\bm{ms}$ & $\bm{man}$ & $\bm{par}$ & $\bm{cc,mj}$ & NA \\
\hline\multicolumn{9}{l}{Transfer Learning from a Single Source - Value Function (Section \ref{sec:tl-value})} \\ \hline
\cite{torrey2006skill} & $\bm{(px,\pi;en_l)}$ & $\bm{obj_S \underrightarrow{h} obj_T}$ & $\bm{ali}$ & $\bm{qf}$ & $\bm{man}$ & $\bm{bias_{\pi}}$ & $\bm{sc}$ & NA \\
\cite{taylor2007cross} & $\bm{(\pi;en_l)}$ & $\bm{S_S \times A_S \underrightarrow{h} S_T \times A_T}$ & $\bm{ali}$ & $\bm{(qf;qf,any)}$ & $\bm{man}$ & $\bm{Q,bias{\pi}}$ & $\bm{sc}$ & NA \\
\cite{taylor2007transfer} & $\bm{(\pi;en_l)}$ & $\bm{S_T \times A_T \underrightarrow{h} S_S \times A_S}$ & $\bm{rel}$ & $\bm{qf}$ & $\bm{man}$ & $\bm{Q,bias_{\pi}}$ & $\bm{sc}$ & NA \\
\cite{taylor2007representation} & $\bm{(\pi;en_l)}$ & $\bm{S_T \times A_T \underrightarrow{h} S_S \times A_S}$ & $\bm{rel}$ & $\bm{qf}$ & $\bm{man}$ & $\bm{Q,bias_{\pi}}$ & $\bm{sc}$ & NA \\
\cite{banerjee2007general} & $\bm{(\pi;en_l)}$ & $\bm{\{S_S,S_T\} \underrightarrow{h} Z}$ & $\bm{app,ali}$ & $\bm{qf}$ & $\bm{man}$ & $\bm{Q,bias_{\pi}}$ & $\bm{bg}$ & NA \\
\cite{kuhlmann2007graph} & $\bm{(\pi;en_l)}$ & $\bm{S_S \underrightarrow{h} S_T}$ & $\bm{ali,app}$ & $\bm{vf}$ & $\bm{man}$ & $\bm{V}$ & $\bm{bg}$ & NA \\
\cite{torrey2008relational} & $\bm{(px;en_l)}$ & $\bm{S_S \times A_S \underrightarrow{h} S_T \times A_T}$ & $\bm{ali}$ & $\bm{qf}$ & $\bm{man}$ & $\bm{Q,\pi}$ & $\bm{sc}$ & \cite{taylor2005valuemethods,torrey2006skill} \\
\cite{taylor2008autonomous} & $\bm{(px;en_l,m)}$ & $\bm{S_T \times A_T \underrightarrow{l} S_S \times A_S}$ & $\bm{none}$ & $\bm{qf}$ & $\bm{man}$ & $\bm{Q,bias_{\pi}}$ & $\bm{cc}$ & NA \\
\hline\multicolumn{9}{l}{Transfer Learning from a Single Source - Bias (Section \ref{sec:tl-bias})} \\ \hline
\cite{torrey2005advice} & $\bm{(\pi;en_l)}$ & $\bm{S_S \times A_S \underleftrightarrow{h} S_T \times A_T }$ & $\bm{ali}$ & $\bm{qf}$ & $\bm{man}$ & $\bm{bias_{\pi}}$ & $\bm{sc}$ & NA \\
\cite{ammar2012commonTL} & $\bm{(s,\pi;s,en_l)}$ & $\bm{S_T \underrightarrow{l} S_S,\: \{S_S,S_T\} \underrightarrow{h} Z }$ & $\bm{ali,rel}$ & $\bm{any}$ & $\bm{man}$ & $\bm{bias_{\pi}}$ & $\bm{cc}$ & NA \\
\cite{wan2020mutual} & $\bm{(\pi;en_l)}$ & $\bm{S_T\underrightarrow{l} S_S}$ & $\bm{rel}$ & $\bm{nn}$ & $\bm{man}$ & $\bm{bias_{par}}$ & $\bm{mj}$ & NA \\
\hline\multicolumn{9}{l}{Transfer Learning from a Single Source - Data (Section \ref{sec:tl-data})} \\ \hline
\cite{taylor2008instances} & $\bm{(px;en_l)}$ & $\bm{S_T \times A_T \underrightarrow{h} S_S \times A_S}$ & $\bm{ali}$ & $\bm{mb}$ & $\bm{man}$ & $\bm{data}$ & $\bm{cc}$ & NA \\
\cite{cao2022learning} & $\bm{(i^l;i)}$ & $\bm{\{S_S \times A_S, S_T \times A_T\} \underrightarrow{l} Z}$ & $\bm{ali}$ & $\bm{bc}$ & $\bm{man}$ & $\bm{ann}$ & $\bm{mj,rp}$ & \cite{zhang2021learning} \\
\hline\multicolumn{9}{l}{Transfer Learning from a Multiple Sources - Sources Combination (Section \ref{sec:tl-combine})} \\ \hline
\cite{ammar2015autonomous} & $\bm{(\pi;en_l)}$ & $\bm{none}$ & $\bm{lin}$ & $\bm{pg}$ & $\bm{multi}$ & $\bm{bias_{par}}$ & $\bm{cc}$ & NA \\
\cite{qian2020intra} & $\bm{(\pi;en_l,or)}$ & $\bm{none}$ & $\bm{lin}$ & $\bm{pg}$ & $\bm{multi}$ & $\bm{bias_{par}}$ & $\bm{cc}$ & \cite{ammar2015autonomous} \\
\cite{heng2022crossdomain} & $\bm{(px,\pi;\nabla en_l)}$ & $\bm{S_S\underleftrightarrow{l} S_T,\:A_S\underrightarrow{l} A_T}$  & $\bm{rel}$ & $\bm{nn}$ & $\bm{multi}$ & $\bm{bias_{par}}$ & $\bm{mj}$ & \cite{wan2020mutual} \\
\cite{liu2023crptpro} & $\bm{(en_r;en_l)}$ & $\bm{none}$ & $\bm{same_O}$ & $\bm{any}$ & $\bm{multi}$ & $\bm{par,shape}$ & $\bm{dm}$ & NA \\
\hline\multicolumn{9}{l}{Transfer Learning from a Multiple Sources - Source Selection (Section \ref{sec:tl-select})} \\ \hline
\cite{talvitie2007experts} & $\bm{(\pi;en_l)}$ & $\bm{S_T \underrightarrow{h} S_S}$ & $\bm{same_A}$ & $\bm{any}$ & $\bm{auto}$ & $\bm{\pi}$ & $\bm{sc}$ & NA \\
\cite{serrano2021inter} & $\bm{(\pi;en_l)}$ & $\bm{none}$ & $\bm{none}$ & $\bm{qf}$ & $\bm{auto}$ & $\bm{Q}$ & $\bm{cc}$ & NA \\
\cite{serrano2023similarity} & $\bm{(px;en_l)}$ & $\bm{S_S \times A_S \underleftrightarrow{l} S_T \times A_T}$ & $\bm{none}$ & $\bm{any}$ & $\bm{auto}$ & $\bm{\pi}$ & $\bm{mj}$ & NA \\
\cite{zhang2024heterotrl} & $\bm{(\pi;en_l)}$ & $\bm{S_T \underrightarrow{h} S_S }$ & $\bm{none}$ & $\bm{ac}$ & $\bm{auto}$ & $\bm{bias_{par}}$ & $\bm{mj}$ & NA \\
\hline
\end{tabular}
\end{table*}

\section{Cross Domain Imitation Learning}\label{sec:il}
This section presents works that learn from expert demonstrations grounded on a domain that is not the learning domain. Section \ref{sec:il-sa} covers methods that use state-action demonstrations, which represent what should be done and in which situation, whereas Section \ref{sec:il-s} describes methods that learn from state-only demonstrations, which capture how the state of the system evolves while an actor performs an expert behavior.

\subsection{Learning from State-Action Demonstrations}\label{sec:il-sa}
Given how difficult it can be to find high-quality sources of knowledge, cross-domain imitation learning methods can exploit expert state-action demonstrations, even if they do not match the target domain. As such, \cite{kim2020domain} learn inter-task mappings between the state and action spaces of a source and target domain to transfer a policy in zero-shot fashion (\textit{i.e.} it is used in the target domain without being finetuned with additional target-domain data), and demonstrations to learn a target policy via a behavior cloning method. Given a set of proxy-task data set pairs (\textit{i.e.} data sampled while learning the same set of tasks in both domains), the mapping functions are trained in an adversarial setting (\textit{i.e.} the mapping models are trained to confuse a discriminator) to preserve the optimality of the source-domain proxy policies translated to their target-domain corresponding task and the transition dynamics across domains. 

Alternatively, \cite{fickinger2021cross} apply an optimal transportation approach to address the data scarcity of cross-domain imitation learning. Instead of learning inter-domain mappings, the Gromov-Wasserstein distance \cite{memoli2011gromov} is used to compare state visitation distributions induced by policies of tasks with different state-action spaces. The divergence between a visitation distribution (approximated) from a single source-domain demonstration and the distributions induced by target-domain policies is used to define an auxiliary reward (\textit{i.e.} by subtracting it from the original reward) that biases the learning policy towards optimal behaviors. That is, to maximize the new reward function, the agent must take actions that produce state visitation distributions that are more similar to the source-domain example.

\subsection{Learning from State Demonstrations}\label{sec:il-s}
Compared to state-action demonstrations, state-only demonstrations are easier to generate, as one must only specify sequences of states/configurations that constitute optimal paths without knowledge of the actions that can cause such transitions. This setting is addressed in \cite{raychaudhuri2021cross} and \cite{salhotra2023learning} by translating expert-generated state trajectories and augmenting them with actions from the target domain. \cite{raychaudhuri2021cross} learn from a set of proxy task data sets (made of unpaired and unaligned trajectories). This inter-domain state mapping maximizes the consistency and distribution matching of states across domains to translate expert demonstrations and augment them with actions via an inverse dynamics model (\textit{i.e.} a model that infers the actions that caused the sequence of states). Alternatively, \cite{salhotra2023learning} take advantage of domains that share the state space to directly add target-domain actions via indirect trajectory optimization \cite{kelly2017introduction}, then feed the state and generated actions into a simulator, which yields state transitions that reflect the target-domain dynamics. Once the simulation-based state demonstrations have been action-augmented, a policy is learned via behavior cloning or an imitation learning algorithm.

Learning mappings across spaces from different domains can be data costly. To improve data efficiency, pretraining models that capture relevant information (about a domain) can yield representations better suited for finding inter-domain correspondences. In \cite{franzmeyer2022learn}, a latent space is learned in the source domain to maximize the mutual information with respect to the state space. Then, a target-domain policy and target-domain state space to latent space mapping are simultaneously trained, where the goal is that a discriminator can not distinguish between source state sequences mapped to the latent space from state sequences visited by the target policy and mapped to the latent space. Similarly, \cite{li2023ace} pretrain a robot-specific (\textit{i.e.} the target domain) motion predictor to infer the next position based on the current position and a latent embedding trained to capture temporal information. Then, a model is trained to map human positions (\textit{i.e.} the source domain) to latent vectors that (when fed to the motion model) produce robot transitions that a discriminator can not differentiate from real robot transitions. Once the mapping is learned, expert demonstrations can be translated to train a target-domain policy via imitation learning.

An alternative to finding inter-domain mappings that are good enough to transfer demonstrations is to learn an embedding/latent space in which states from different domains can be compared for reward-shaping purposes. For instance, \cite{zakka2022xirl} fine-tune an Imagenet-trained Resnet model \cite{targ2016resnet,deng2009imagenet,russakovsky2015imagenet} on a set of different robot demonstrations to be temporally aligned \cite{dwibedi2019temporal}, and use it to map images into embeddings (\textit{i.e.} vectors from the latent space). Then, the embedding of the goal state of every demonstration is averaged and used as a reference for the embedding of the current state of the target-domain learner. Via reward shaping, the learner policy is encouraged to move towards states that are closer (in the latent space) to the average goal state. On the other hand, \cite{li2023crossloco} train a pair of human-robot (\textit{i.e.} source and target domains) mappings to preserve consistency of their positions through the mappings, while the robot policy (conditioned on the human position) is simultaneously trained via reward shaping to perform actions that transition into positions, whose human equivalent (according to the inter-domain mapping) is close to the actual human next position.

\section{Cross Domain Transfer Learning}\label{sec:tl}
In contrast to imitation learning (see Section \ref{sec:il}), which learns from expert demonstrations, transfer learning methods strive to exploit knowledge discovered in a previous task to aid learning better in the current task. This section covers such works, organized as follows: Section \ref{sec:tl-single} covers methods that transfer from a single source of knowledge and are classified based on the type of knowledge being transferred. Section \ref{sec:tl-multi} presents works focused on transferring from a set of knowledge sources, whether multiple sources are combined (Section \ref{sec:tl-combine}) or knowledge is transferred from the best source of knowledge (Section \ref{sec:tl-select}).

\subsection{Single-Source Transfer}\label{sec:tl-single}
\subsubsection{Demonstrations}\label{sec:tl-demos}
Representing knowledge as demonstrations allows the use of a variety of learning algorithms since no data-format restrictions are imposed. In some cases, transferring demonstrations can help compute a good parameter initialization without interacting with the target environment. In \cite{ammar2012sparsecoding}, a state-action inter-domain mapping is learned to transfer state transitions from the source domain that initialize a target-domain learner. After learning a linear decomposition that approximates a source state-transition data set (\textit{i.e.} $\langle s_{i},a_{i},s_{i+1} \rangle$) \cite{lee2006efficient}, one of the learned factors is used to learn a projection of a target state-transition data set to the same latent space. Then, every target tuple is paired to the closest source tuple (according to their latent projections) and used to train a regressor as an inter-domain mapping function. To transfer knowledge, a source optimal policy greedily samples state-transition tuples and maps them to the target domain, which the learning policy uses to compute an initial evaluation and parameter update. Similarly, \cite{ammar2015unsupervised} learns an inter-domain state mapping \cite{wang2009manifold} to map a set of initial states (from trajectories sampled in the target domain) to the source domain. Then, roll out an expert source police (starting on the mapped initial states) and map the expert state trajectories to the target domain so that the policy has a better parameter initialization by minimizing a mimic-behavior loss (with respect to the mapped trajectories).

Learning a latent space of skills or strategies helps to separate task-invariant information from domain-dependent features, which leads to better generalization across domains. For instance, \cite{shankar2020learning} learns a skill-tuple space for each domain (which has temporal information extracted from state-action sequences), then learns a source-to-target mapping that aligns the skill-tuple distributions, which allows transferring state-action policy demonstrations (in zero-shot fashion), or fine-tune them to learn a target-domain policy. In the case of \cite{aktas2023correspondence} and \cite{watahiki2023leveraging}, from a set of proxy tasks, a single and shared latent policy is learned, which can receive states and generate actions from any domain. In \cite{aktas2023correspondence}, a set of robot-specific models \cite{seker2019conditional} is learned jointly, which allows generating actions for any robot given observations sampled from successful trajectories of a different robot. On the other hand, \cite{watahiki2023leveraging} trains an encoder-decoder pair along with a latent policy that can act in every domain. Then, the encoder and decoder models are frozen, and the policy is fine-tuned to mimic the behavior captured in a set of source-domain demonstrations. Finally, the policy is evaluated in the target domain without further tuning.

\subsubsection{Policy}\label{sec:tl-policy}
To transfer policies across domains, it is necessary to adapt them to the target domain; such process varies depending on how different the domains are. For instance, \cite{cheng2018reusing,joshi2018cross,chen2024mirage} take advantage of settings where the action space is shared across domains. Given a user-provided state mapping, \cite{cheng2018reusing} maps target states to evaluate their action value for every source action, according to a trained source Q function (\textit{i.e.} the action-value function) \cite{watkins1992q}. Then, a model (trained to assess the state-action quality) is used to rank multiple state-action pairs and transfer the best action to be performed in the target domain, given a certain probability. In \cite{joshi2018cross}, a state mapping is learned \cite{wang2009manifold} and used to transfer a source policy to the target domain, which is biased with the addition of an adjustment term, defined as the difference between the next-state predictions made by the transition models of both domains. The adjustment term adapts the policy to correct differences between the domains' transition dynamics.

Moreover, to avoid learning state mappings, \cite{chen2024mirage} assume the availability of robot simulators for both domains and their respective URDF (\textit{i.e.} unified robot description format \cite{tola2023understanding}, which completely describes the robot's kinematics). They propose to erase the target robot from an image, render the source robot in the same position, and paint it in. Then, the altered image is fed to an already trained source policy, which outputs an action that is executed by the target robot, and the cycle repeats.

To relax the shared-space assumption, \cite{soni2006homomorphism} use a set of one-to-one mappings the user provides to transfer policies. Given a set of options \cite{sutton1998intra} (\textit{i.e.} abstract/temporally extended actions) learned in the source domain, \cite{soni2006homomorphism} propose evaluating in which regions of the target-domain state space each option (transferred with one of the provided mappings) is best suited. The option-fitness criterion is based on the definition of MDP homomorphisms \cite{ravindran2004algebraic,sorg2009transfer} and focuses on preserving the alignment of the transition dynamics while leaving out the reward constraints from the original definition. In addition, an action-value function is simultaneously learned to use the transferred options. That is, the proposed method searches in different regions of the target state space for the mapping that better preserves state consistency (given the transition dynamics) while learning an action-value function to use the mapped options to solve the task in the target domain.

Similarly, \cite{wang2022weakly} use a weak form of supervision to find correspondences between domains. In addition to using proxy data sets, \cite{wang2022weakly} exploit user-provided soft cross-domain pairings of states and state-action pairs, in which a real number describes how similar they are. Combining the weak supervision, in the form of similarity scores, with a multi-step transition dynamics consistency loss (which encourages the mapping to align states and actions with respect to the transition functions), a state-action mapping is learned and used to transfer a source policy to the target domain.

Regarding works that learn inter-domain mappings with less supervision, both \cite{zhang2021learning} and \cite{gui2023cross} focus on learning state-action mappings that align the transition dynamics across domains to transfer a source policy to the target domain for immediate usage (\textit{i.e.} zero-shot transfer). On the other hand, after learning a latent action embedding for each domain \cite{pertsch2021accelerating}, \cite{yang2023learn} learn a pair of mappings across the individual latent spaces, which must preserve consistency and align distributions, using paired proxy data sets as a reference of which action distributions correspond to the same task. Then, to initialize a target-domain policy for an unseen task, the source-domain prior skill for that task is transferred through the latent-space mapping.

Similarly, \cite{shoeleh2020skill} propose learning a latent shared space in which experience tuples $\langle s_i,a_i,s_{i+1},r_i \rangle$ sampled from a source and target domain can be compared based on their similarity, using a domain adaptation method \cite{pan2010domain}. Then, the latent space is used to supervise the training process of a pair of mappings that transform target-domain state and actions to the source domain space, which are used to finetune a set of source options (\textit{i.e.} abstract actions) \cite{sutton1998intra,sutton1999between} to maximize the returns in the target domain. Additionally, \cite{shoeleh2020skill} propose transferring the action values of the source options (through the learned latent space) to serve as the supervision signal in an offline training process of options for the target domain task.

\subsubsection{Reward Shaping}\label{sec:tl-shape}
The main idea behind reward shaping is to communicate a learning algorithm (through an auxiliary reward) how similar it behaves to an optimal policy. This behavioral similarity can be measured directly \cite{brys2015policy}, as the probability assigned by a source policy to a target state-action mapped to the source domain. This probability is used to learn a secondary value function (\textit{i.e.} potential function) that behaves as the auxiliary reward for the target domain policy. Moreover, an important reason to use a potential-based approach is that it guarantees that the auxiliary reward will not change the task induced by the original reward \cite{ng1999policy}. That is, a policy that maximizes the shaped reward function is also optimal with respect to the original reward.

Conversely, if the similarity to an optimal policy can not be directly evaluated, elements associated with policies' behavior can be used to measure similarity indirectly. For instance, from a set of paired and aligned demonstrations, \cite{gupta2017learning} and \cite{hu2019skill} learn a shared latent state/state-action space to compare states/state-action pairs visited by a source expert and a target learner policy, with the computation of the Euclidean distance/Kullback-Leibler divergence (\textit{i.e.} a distance function for probability distributions) between them. Similarly, \cite{hejna2020hierarchically} define an auxiliary reward for low-level policies in a hierarchical framework. Given a high-level state variable (\textit{e.g.} the global position of a robot in a maze), the target domain low-level policies are encouraged to behave as their source domain counterpart while reaching the same subgoal. The indirect similarity measure (\textit{i.e.} auxiliary reward) is defined by the score a discriminator (trained to identify source low-level behaviors) assigns to the target-domain low-level policies, using the value of the high-level state variable as input. That is, the auxiliary reward is maximized by visiting states that confuse the discriminator into classifying them as states visited by the source policy.

\subsubsection{Parameters}\label{sec:tl-params}
When the learned solution is represented by the same type of model in the source and target domain (\textit{e.g.} using a neural network in both domains), transferring knowledge in the form of model parameters provides excellent control over what is being transferred, and what is being left to be learned in the target domain. For instance, \cite{taylor2005valuemethods,taylor2005behavior,taylor2007tvitmps} use state-action mappings (provided by an oracle/external entity) to map target-domain state-action pairs to the source domain. Then, they identify the parameter associated with each state-action pair and transfer it to its counterpart in the target-domain model. The transferred parameters initialize the model, which will finish its refinement through a standard RL process.

On the other hand, \cite{devin2017learning} propose a transfer framework in which domains are defined by two independent elements: a task (\textit{e.g.} pushing objects, opening a drawer) and an agent (\textit{e.g.} a robot arms with different degrees of freedom). A policy for a particular domain is made of a task-dependent network, whose output is fed (along with agent-dependent state variables) into an agent-specific network that yields the actions performed to solve a task. After training on a combination of task-agent pairs (assumed to include every task and agent, while not every combination), a new unseen task-agent domain is solved (in a zero-shot fashion) by transferring the task and agent networks to build a domain-specific policy.

Similarly to \cite{devin2017learning}, \cite{chen2019learning} and \cite{zhang2021feature} transfer neural network weights to initialize a policy for the target domain. In \cite{chen2019learning}, the weights of an approximated transition model (which was trained in tandem with an action embedding that allows using SAC \cite{haarnoja2018soft} in tasks with discrete action spaces) and the middle layers of a policy network are transferred. After initializing the target transition model and policy with the transferred weights, they are refined (in addition to the target-domain action embedding) by interacting with the target environment. In the case of \cite{zhang2021feature}, an architecture of multiple subnetworks (each of which represents an option \cite{sutton1998intra}), whose interactions with the environment are handled by a pair of input and output layers (trained for that specific environment), is trained in a source domain before transferring the option subnetworks to initialize the target-domain model. The input and output layers are trained from scratch in the target domain to match the target state and action space dimensions to use the transferred subnetworks. This way, the parameters orchestrating high-level strategies are reused (\textit{i.e.} the option subnetworks). In contrast, the domain-dependent parameters are learned from scratch (\textit{i.e.} the input and output layers).

\subsubsection{Value Function}\label{sec:tl-value}
Transferring value functions is a particular form of parameter transfer, which contain a significant amount of knowledge (\textit{i.e.} the expected return from a state or state-action pair). Furthermore, action-value functions have the flexibility of acting as a policy using a maximization operation with respect to the action space. For example, transferring state/action values to initialize the target domain model \cite{taylor2007representation,banerjee2007general,kuhlmann2007graph,torrey2008relational} can provide an advantage in the learning process when compared to a random initialization. While in \cite{taylor2007representation}, a user-provided state-action mapping is used to transfer source action values to the target domain and use them to train an action value predictor that acts as an initial action-value function; \cite{torrey2008relational} propose modeling knowledge as relational macros and learning a list of rules (with an inductive learner approach) from a set of source-domain $\langle s_{i},a_{i},r_{i} \rangle$ tuples, to achieve successful runs. Then, the rules are translated to the target domain (via the state-action mappings provided by the user) and executed for a fixed number of episodes (in which the action value is the success score associated with the rule being followed). The pseudo action values partially initialize the target model and are progressively removed as the learning process continues. In this way, alternative success metrics improve the initialization of the action-value function compared to randomly assigned values.

Similarly, \cite{banerjee2007general} and \cite{kuhlmann2007graph} transfer values to initialize the target-domain model for turn-taking games. In both works, a game-invariant representation is used to represent state features. Every feature is a tree structure that describes several possible outcomes and for which the agent learns an associated value. After learning the value for a set of features in the source domain, \cite{banerjee2007general} seeks target-domain features that match a source-domain feature to transfer the value to initialize the target model. Moreover, \cite{kuhlmann2007graph} evaluates the overall structure of different games by modeling them as rule graphs (which represent the critical aspects of games). After building the rule graph of a source game, multiple modifications are made to generate a set of transformed graphs that will be compared to future games. If a match is made (by checking for isomorphism) between a new game and one of the transformed graphs (including the original), then values are transferred for matching features similarly to \cite{banerjee2007general}.

In addition to initializing the target-domain value function, source-domain value functions can bias the learning process of the target agent for an extended period of time. In \cite{taylor2007cross}, \cite{taylor2007transfer} and \cite{torrey2006skill}, a user-provided state-action/object mapping is used to transfer action values to accelerate the target-domain learning process. In \cite{taylor2007cross} the source-domain action value of a source state-action pair $\langle 
s_S,a_S \rangle$ is added to its target-domain equivalent $\langle 
s_T,a_T \rangle$ (according to mappings) if $a_S$ is the preferred action from state $s_S$. Additionally, a policy transfer approach is evaluated. While \cite{taylor2007transfer} computes the sum of the action value of every target-domain state-action pair and its source-mapped state-action pair to bias the target model learning process, \cite{torrey2006skill} intervenes a linear optimization process of the target Q function, by imposing a linear restriction with the action value of the action suggested by a set of logic rules (learned from the source Q function and translated with inter-domain objects mapping). Similarly, \cite{taylor2008autonomous} also bias the target learner by adding the source action values to their target equivalent. However, \cite{taylor2008autonomous} proposes to learn the state-action mappings from scratch, which decreases the supervision required to transfer knowledge.

\subsubsection{Bias}\label{sec:tl-bias}
Considering the many factors that affect the reinforcement learning process, the inductive bias applied by transfer knowledge methods may not fit into the most conventional forms of knowledge (\textit{e.g.} policy, parameters, reward shaping). For instance, \cite{torrey2005advice} transfer \textit{if-then} rules (learned from the source-domain Q function) to influence the action-value learning process in the target domain. Given a user-provided state-action mapping that translates rules to the target domain, rules follow the form: \textit{if an action is preferred over all others according to the source Q function, then its target counterpart should be preferred too}. Thus, this decision list behaves as a meta-policy that suggests actions that are \textit{significantly} better than the other ones available. 

In \cite{ammar2012commonTL}, a criterion to select data (later used to train an initial policy) is transferred to the target domain. Given a pair of mappings that transform states from both domains to a common space and a source expert policy, an inter-domain state mapping $f:S^{target} \rightarrow S^{source}$ is learned (via locally weighted regression \cite{atkeson1997locally}). After sampling source state transitions with the source expert policy $\pi^\ast$ (\textit{i.e.} $\langle s^{source}_i,a^{source}_i,s^{source}_{i+1} \rangle$ where $a^{source}_i\triangleq\pi^\ast(s^{source}_i)$), a training set of target state transitions $\{ \langle s^{target}_i,a^{target}_i,s^{target}_{i+1} \rangle \: | \: f(s^{target}_i) = s^{source}_i, d(f(s^{target}_{i+1}),s^{source}_{i+1}) < \epsilon\: \}$ is built, where $d(\cdot,\cdot)$ is a distance function. That is, the target-domain data set will consist of state transitions whose initial state $s^{target}_i$ and its source-domain equivalent $f(s^{target}_i)$ can transit to states $s^{target}_{i+1}$ and $s^{source}_{i+1}$ (respectively), such that the source next-state $s^{source}_{i+1}$ and the target next-state mapped to the source domain $f(s^{target}_{i+1})$ are close enough. Then, the generated training set is used to train an initial policy that will hopefully reduce the exploration required to solve the target task.

As an alternative bias for neural network models, \cite{wan2020mutual} transfer the preactivation output of hidden layers (\textit{i.e.} the sum of the weighted inputs to a node) to linearly combine it with the preactivation output (of the correspondent layer) of a student network (\textit{i.e.} the policy in the target domain). To handle the state mismatch between the source and target domains, a mapping is learned to map target states to the source domain, which is encouraged to maximize the mutual information across state spaces. To transfer knowledge, a target domain state and its source domain embedding are fed into the student and teacher models, respectively. Then, the preactivation sum of each hidden layer is linearly combined (and weighted by a parameter that is adjusted during the training process) with the preactivation of the equivalent layer in the teacher model. The combination weights, state mapping, and student model are all trained simultaneously.

\subsubsection{Data}\label{sec:tl-data}
When an approximation of the environment's dynamics is required before learning the policy, transferring raw data can improve the estimation, regardless of the policy used to sample it. For instance, \cite{taylor2008instances} use an inter-domain state-action mapping to translate source instances $ 
\langle s^{source}_i,a^{source}_i,s^{source}_{i+1},r^{source}_i\rangle$ to the target domain when a model-based \cite{moerland2023model} method has insufficient data to approximate the reward/transition model for a target state-action pair.

On the other hand, \cite{cao2022learning} propose to transfer annotations of a source-domain data set. The annotations describe the confidence of a demonstration having a good quality (\textit{i.e.} expert-generated sequence of states and actions), and by transferring them to a target-domain data set, an imitation learning method can learn from data sets containing imperfect demonstrations. After learning a latent embedding to which source state-actions are mapped and a decoder $F$ that maps embedding vectors to confidence scores (supervised by the source annotations), a target encoder (\textit{i.e.} maps target state-actions to the latent embedding space) is trained to confuse a set of discriminators so they can not distinguish between the embeddings and scores assigned by $F$, of state-action sequences from both domains. Moreover, a multi-step alignment loss allows finding correspondences without paired trajectories. Once the target encoder is trained, the latent embedding of every state-action pair in the target-domain data set is fed into $F$ to annotate them with a confidence score that a behavior cloning method can use to weight the importance/quality of each demonstration.

\subsection{Multiple-Source Transfer}\label{sec:tl-multi}

\subsubsection{Source Combination}\label{sec:tl-combine}
In a successful multi-source knowledge combination, the different pieces of information complement each other, thus producing a robust model for various situations. Some works synthesize multiple sources of knowledge into a single model before the transfer process begins. For instance, \cite{liu2023crptpro} samples (with random policies) observations from multiple domains and use them to train an image encoder and a set of latent vectors (\textit{i.e.} prototypes) that serve as cluster centers. During training, in a self-supervised manner, the image encoder is trained to predict (with the current observation as input) the cluster assignment of the next observation. In contrast, the prototypes are encouraged to increase their coverage of the latent space. When a new task is encountered (from an unseen domain), the RL algorithm uses the representation yielded by the image encoder instead of the original image, and an auxiliary reward promotes actions that transit observations that are near (in the latent space) to one of the learned prototypes.

Conversely, other works incorporate feedback from the target-domain learning process to decide how to combine the knowledge. In  \cite{ammar2015autonomous}, a cross-domain variation of the lifelong learning setting \cite{parisi2019continualsurvey,khetarpal2022continualsurvey}, is addressed with a hierarchical knowledge representation. Task-specific policy parameters are modeled as a linear combination of a latent basis shared across domains, a domain-specific basis shared across tasks (from the same domain but with different dynamics), and a task-specific vector. Every time a new task is encountered, the shared basis and domain-specific bases are updated based on the observations that have been recently gathered. Then, the task-specific coefficients are updated with respect to the most recent version of the shared and domain-specific bases. This interleaved process allows updating the shared knowledge with every new observation, as well as to transfer the shared knowledge to improve the task-specific policy. Moreover, \cite{qian2020intra} expands on the framework proposed by \cite{ammar2015autonomous} to improve the intra-domain generalization when coefficients that determine the environment dynamics can be observed by including them in the computation of the domain-specific basis.

Similarly, \cite{heng2022crossdomain} uses target-domain feedback to determine how the set of knowledge sources will be combined for transfer purposes. Given a set of source policies (modeled as neural networks), \cite{heng2022crossdomain} learns (during the target-domain task training) an inter-domain state and action mapping between the target domain and each source domain. The mappings strive to maximize the mutual information, consistency, and alignment of the transition dynamics between the source and target domain, as well as to maximize the target policy performance. Afterward, the mappings are used to evaluate the performance of each source policy in the target domain. Then, the performance scores are normalized to weight the influence of each source policy via hidden-layer preactivation outputs, in addition to a coefficient that controls the source policies' overall influence (similarly to \cite{wan2020mutual}). The mapping updates, evaluation of source policies, and weighted transfer to the learning policy are repeated throughout the training process.

\subsubsection{Source Selection}\label{sec:tl-select}
In contrast to source-combining approaches, selecting a single source of knowledge among multiple sources can be very helpful when uncertain whether every source candidate is related to the target domain. Thus, before transferring knowledge, the compatibility between the target domain and every source domain is evaluated to find the best option. Some works measure the inter-domain similarity once, at an early stage of the learning process, to transfer as soon as possible. For instance, \cite{serrano2021inter} measures the similarity between two domains by comparing their action-value functions. The action values associated with every state and action (separately) are clustered, and the assignment distributions are compared against states and actions from the other domain (using histogram intersection). The cross-domain state and action similarity values are stored in two matrices whose norms are used as similarity scores. Then, action values are transferred to the target model from the most similar source domain.

Similarly, \cite{serrano2023similarity} uses the first interactions (sampled by the target learning agent) to learn an inter-domain state-action mapping between the target domain and every source domain, following a reward-based alignment criterion. Then, the mappings are used to compare the transition and reward dynamics (using approximated models). The policy from the most similar source domain is transferred (with the inter-domain mappings) for a fixed number of episodes to alleviate the exploration process.

On the other hand, some methods evaluate the inter-domain similarity throughout the target learning process to update the selected source according to the most recent observations. In \cite{talvitie2007experts}, with the help of user-provided state-action mappings, every source domain expert policy is evaluated in the target domain (for a fixed number of episodes), and their returns are stored. Then, the policy with the highest return is used for an extended period of steps. If, at any point in time, the selected policy obtains lower returns than its test score, it is eliminated from the set of available experts, and the next-best policy is transferred. When the set of available experts is empty, the evaluation-transfer cycle repeats. Likewise, \cite{zhang2024heterotrl} implement a two-fold cycle in which similarity between the target domain and every source domain is evaluated, followed by transferring knowledge from the best option. In this case, inter-domain similarity is computed as the difference in the state values generated by the target and each source critic (assuming agents are modeled with an actor-critic architecture) for a sequence of states sampled in the target domain. The source model with the lowest difference is selected as the knowledge source, whose advantage function will be added to the target domain advantage to bias the target policy. The source selection and knowledge transfer process is repeated until the target task training ends.

\section{Discussion}\label{sec:discussion}
This section presents a series of open questions regarding knowledge transfer in cross-domain RL that remain to be further studied, as well as some ideas to explore in the future.

\subsection{Negative Transfer and Autonomous Knowledge Transfer}
Negative transfer, which refers to the phenomenon of harming the performance of a learning agent as a consequence of using auxiliary knowledge, has been progressively mitigated with the development of methods that use information about the target domain to decide how to transfer knowledge \cite{taylor2008autonomous,ammar2015autonomous,heng2022crossdomain,serrano2023similarity,zhang2024heterotrl}. These methods focus on a particular form of negative transfer: to identify the best source of knowledge from a library of good and bad options. However, to the best of our knowledge, there is no evidence of methods that successfully identify that none of the available sources of knowledge will improve the target learner.

The main challenge in avoiding negative transfer in RL is deciding whether a source of knowledge can improve a target learner's performance (through knowledge transfer) while consuming less data than the one required to reach threshold performance by learning from scratch. Fortunately, in the same way that there is no general definition of inter-MDP similarity for transfer purposes \cite{garcia2022taxonomy}, evaluating inter-domain compatibility should be focused on attributes related to the type of knowledge being transferred, as the negative transfer verification is dependent on the knowledge transfer method.

Whether it is by comparing transition dynamics \cite{ammar2014automated,castro2020scalable}, reward dynamics \cite{carroll2005task,gleave2020quantifying,tao2021repaint}, or both \cite{ravindran2004algebraic,sorg2009transfer,castro2010using,castro2011automatic,song2016measuring,serrano2023similarity}, MDP similarity measures can characterize domains/tasks with enough detail to determine if positive transfer is possible within a pair of tasks. However, to address the negative transfer problem, we believe that using methods to measure uncertainty (\textit{e.g.} Gaussian processes \cite{rasmussen2003gaussian,bilionis2012multi}, statistical significance \cite{dangeti2017statistics}) could improve the selection of sources of knowledge by incorporating the confidence of the inter-MDP similarity measurements.

For instance, transfer learning methods endowed with mechanisms to select the \textit{best} source of knowledge \cite{talvitie2007experts,serrano2021inter,serrano2023similarity,zhang2024heterotrl} could benefit from an uncertainty measuring method. Similar to how fitness/compatibility is measured in \cite{talvitie2007experts}, by evaluating source policies in the target domain, one could have a way to compare whether any of the sources of knowledge can improve the target policy. If the target policy, in its partially learned form, outperforms the best source policy, then one could choose to learn from scratch.

\subsection{Formalization of Inter-Domain Similarity Assumptions}
Satisfying the method's assumptions is a crucial step in obtaining successful results before using a transfer method. Whether it is sharing the state, action, or observation space or having a similar high-level structure, many of the reviewed works make assumptions about source and target domains being similar in some way. However, in the case of similar high-level/latent structures, it becomes unclear if a setting (different from the one used in the original publication for evaluation) meets this criterion.

Formalizing the conditions under which cross-domain alignment is beneficial for knowledge transfer purposes, along with developments in theoretical work, will make finding the right method for a particular problem easier. For instance, with the introduction of \textit{\textbf{MDP reduction}}, \cite{kim2020domain} provides a formal description of the characteristics that a pair of MDPs must have in order to transfer optimal policies that will preserve an optimal behavior in the target domain. Similarly, \cite{soni2006homomorphism} uses a relaxed definition of MDP homomorphism \cite{ravindran2004algebraic} to select the best inter-domain state mapping to transfer policies. Thus, further research in theoretical work and data-efficient approximations are critical to improve the robustness of knowledge transfer methods in the cross-domain setting.

\subsection{Too Expensive for Real-World Applications}\label{sec:disc:too-expensive}
As with many RL subareas, cross-domain knowledge transfer is still too data-costly for most real-world applications. Among the reviewed works, only a few showed results on real robots. Given that the primary motivation of these methods is to widen the range of scenarios in which knowledge can be reused, it is necessary to consider the needs of real-world domains. Some ideas to tackle the sample complexity of real-world robots are to combine sim-to-real approaches \cite{salvato2021sim2realsurvey} with a refinement/adaptation phase of policies pretrained in simulation or extensive data sets \cite{depierre2018jacquard,dasari2019robonet,walke2023bridgedata,padalkar2023open,khazatsky2024droid}. For instance, by carefully selecting a sequence of simulated tasks in which knowledge is learned and transferred to the next task, knowledge acquired in simulation can significantly improve both the learning speed and final performance in comparison to learning from scratch with the physical robot \cite{barrett2010transfer}.

Additionally, many situations encompassing robots performing physical movements (\textit{e.g.} navigating, manipulating objects, walking while keeping balance) to assist in human-related tasks are not episodic. Tasks with long horizons can be particularly challenging for standard RL methods \cite{mnih2015human,pateria2021hierarchicalsurvey}. Hierarchical-based knowledge transfer provides an alternative to learning in long-horizon tasks because of the advantage of performing temporal abstraction (through temporally extended actions) when exploring for extended periods of time. Therefore, combining strategies that handle inter-domain mismatches with hierarchical models would significantly extend the settings where knowledge can be reused.

For instance, \cite{hejna2020hierarchically} transfer behavior across agents with different morphologies by decoupling high-level and low-level behaviors. By doing so, if the overall strategy remains the same across domains (\textit{i.e.} high-level behavior), only the low-level policies need to be fine-tuned to fit the target agent's physical structure, which might help reduce the data required to learn the real-world task. Additionally, besides decoupling behaviors at different levels of granularity, exploiting knowledge about the simulated agent and its non-virtual counterpart could mitigate the number of required real-world samples. Similar to how \cite{wang2022weakly} use state-action pairings between the source and target agent, one could provide this type of weaker supervision to help correct the error in the simulated dynamics via an inter-domain mapping so that the policy trained in simulation can be mapped and used in the real world, without any further tuning.

\subsection{Practical Implications and Suggestions}
Although some environments are considered to be solved by standard RL algorithms (\textit{e.g.} taxi domain \cite{dietterich2000hierarchical}), knowledge transfer methods are still too data-expensive to be used in most real-world problems, as discussed in Section \ref{sec:disc:too-expensive}. However, given how time-consuming training RL agents can be, even in simulated environments, knowledge transfer strategies have an opportunity to make an immediate impact on the development of robotics learning, whether it is with imitation learning or transfer learning for industrial, medical or service robotics \cite{hua2021learning}. Through transfer and imitation learning methods, RL agents can be trained in less time, thus presenting an alternative to decrease the time required to test ideas without using graphic processing units (GPU).

Regarding practical tools, multiple frameworks make starting as an RL practitioner easier. Table \ref{tab:rl-libraries} presents a list of frameworks (all free to use) that facilitate implementing knowledge transfer ideas without coding everything from scratch. On the other hand, Table \ref{tab:rl-simulators} shows a list of simulators that include a wide variety of sequential decision-making problems. Moreover, given that every framework from Table \ref{tab:rl-libraries} and Table \ref{tab:rl-simulators} is an independent project, one can couple the pair of RL packages and simulator that better fits the available computational resources.

Additionally, it is a common practice for RL packages/libraries to provide benchmark results of standard RL algorithms (\textit{e.g.} SAC \cite{haarnoja2018soft}, PPO \cite{schulman2017proximal}) on popular problems. Using these evaluation results (as baseline methods) can save training time that could be better employed in testing original ideas, or tuning the model's parameters. In online RL, given that the learning agent is partially responsible for generating the training data (\textit{i.e.} due to its interactions with the environment), it can be challenging to predict the effect of changing parameter values. Thus, tuning the model hyperparameters can take multiple iterations.

\begin{table}[t]
\caption{Reinforcement Learning packages/libraries. Considering that a significant amount of machine learning systems are coded in Python, for compatibility reasons, the second column shows the underlying framework used by the package: PyTorch \cite{paszke2017automatic}, TensorFlow \cite{tensorflow2015-whitepaper}, or it is compatible with both frameworks.}
\label{tab:rl-libraries}
\centering
\begin{tabular}{ll}
\hline
\multicolumn{1}{c}{RL Package} & \multicolumn{1}{c}{Underlying Framework} \\ \hline
CleanRL \cite{huang2022cleanrl} & PyTorch \\
Coach \cite{caspi_itai_2017_1134899} & TensorFlow \\
Dopamine \cite{castro2018dopamine} & TensorFlow \\
Mushroom RL \cite{deramo2021mushroomrl} & Both \\
ReAgent \cite{gauci2018horizon} & PyTorch \\
RL Lib \cite{liang2018rllib} & Both \\
Stable Baselines3 \cite{stable-baselines3} & PyTorch \\
TensorFlow Agents \cite{tfAgents} & TensorFlow \\
MLPro \cite{arend2022mlpro} & PyTorch \\ \hline
\end{tabular}
\end{table}

\begin{table}[t]
\caption{Simulators of multiple types of problems that can be used as environments in a reinforcement learning setting/experiment. Simulators are independent of the RL library and, thus, can be easily coupled with any of the frameworks in Table \ref{tab:rl-libraries}.}
\label{tab:rl-simulators}
\centering
\begin{tabular}{ll}
\hline
\multicolumn{1}{c}{Simulator} & \multicolumn{1}{c}{Type of Environments} \\ \hline
Arcade Learning Environment \cite{bellemare13arcade} & Atari 2600 games \\
Deepmind Control Suite \cite{tunyasuvunakool2020} & Physics-based control \\
Gymnasium \cite{towers_gymnasium_2023} & Toy text, control, atari \\
Issac Gym \cite{makoviychuk2021isaac} & \begin{tabular}[c]{@{}l@{}}GPU-optimized physics-based\\ control\end{tabular} \\
Malmo \cite{johnson2016malmo} & Minecraft tasks \\
PyBullet \cite{coumans2020pybullet} & Physics-based control \\
Robosuite \cite{zhu2020robosuite} & Physics-based robot learning \\
MLPro-MPPS \cite{yuwono2023mlpro} & Configurable production systems \\ \hline
\end{tabular}
\end{table}

\subsection{On-Policy and Off-Policy Learning in Knowledge Transfer}
One of the most critical differences RL algorithms can have is whether the policy being learned (\textit{i.e.} target policy) is the same one responsible for generating actions (\textit{i.e.} behavior policy) to explore the environment (\textit{i.e.} on-policy learning), examples of such methods include SARSA \cite{rummery1994line} and Policy Iteration \cite{howard1960dynamic}. On the other hand, methods that use two policies (independent from each other) to carry the learning and exploration processes are known as off-policy learning methods \cite{sutton2018reinforcement} (\textit{e.g.} Q-learning \cite{watkins1992q}, Expected SARSA \cite{van2009theoretical}). It is worth mentioning that both on-policy and off-policy methods can be used in the online learning setting (\textit{i.e.} the agent gathers data by interacting with the environment), while only off-policy methods can be used in the offline learning scenario since the data set was generated with a different policy than the one that will perform the learning updates.

Generally, off-policy methods tend to be more sample efficient since the behavior and target policies can be designed separately to specialize on exploring the environment and learn as efficient as possible, respectively, without compromising each other's objective\cite{gu2016q}. Additionally, off-policy methods are considered to be more flexible than their on-policy counterpart given that they can learn with data that was sampled by multiple actors/policies, as the target policy does not need to be the one generating the actions responsible for the exploration.

Moreover, the flexibility of off-policy learning methods extends to the particular case of knowledge transfer, as 43\% of the reviewed works (23 out of 53) transfer knowledge in a form that only an off-policy algorithm could use in the target domain (\textit{i.e.} policies, demonstrations and samples from interactions with the source environment). However, on-policy methods can offer better solutions when online performance is critical (\textit{i.e.} the behavior policy quickly reflects the effect of learning updates) \cite{perkins2002convergent}, or low computational complexity restrictions must be met, as only one policy needs to be stored and updated \cite{fakoor2020p3o}. In such scenarios, transferring knowledge in the form of reward shaping \cite{brys2015policy,gupta2017learning,hu2019skill,hejna2020hierarchically,fickinger2021cross,zakka2022xirl,li2023crossloco}, or model parameters \cite{taylor2005valuemethods,taylor2005behavior,taylor2007tvitmps,devin2017learning,chen2019learning,zhang2021feature,liu2023crptpro}, can provide a powerful alternative to improve its sample efficiency without generating off-policy data.

\subsection{Challenges in Results Reproducibility}
A more practical concern, but equally important, is the set of challenges encountered when trying to reproduce an experiment in knowledge transfer for the RL setting. A reproducibility checklist for machine learning works is proposed in \cite{pineau2021improving} (a report on a reproducibility program for the NeurIPS 2019 conference). The checklist requires a complete description of the models used (mathematical model and parameters), proofs and assumptions of theoretical claims, the methodology employed to gather data, statistical tools, and links to code, data sets, or simulated environments. With the proposed checklist, the program's authors expected to encourage other authors to improve the transparency of their methodology and results.

However, some issues prevail regarding the reproducibility of the publication of knowledge transfer for RL results, which include:
\begin{enumerate}
    \item \textbf{Disappearance of Online Repositories}: With access to free online repositories, sharing code implementations is easier than ever. Unfortunately, repositories that go missing after the correspondent method has been published are common. Moreover, since works provide access to their code, the implementation details are considered redundant and usually omitted in the article.

    \item \textbf{Ambiguous Data Recollection}: Contrary to supervised and unsupervised learning, where data sets are static, the observed data in RL depends on the constantly evolving agent/policy. Due to the availability of the most used simulated environments (\textit{e.g.} Mujoco \cite{todorov2012mujoco}, Deepmind control suite \cite{tassa2018deepmind}), providing the exact data used in experiments is an unnecessary practice. However, describing the policy responsible for sampling the different data sets is often overlooked. Using different policies to sample data may significantly impact experiments, as this data also influences the data being observed in the future.
\end{enumerate}

Therefore, we suggest providing complete descriptions of the proposed solutions and a detailed explanation of the methods for gathering data, whether it is a random or a trained policy. In the latter case, detailing under which circumstances the policy was trained (\textit{e.g.} the number of environment steps, learning algorithm, learning parameters, learning environment) is crucial to reproduce restults.

\section{Final Remarks}\label{sec:final-remarks}
This article presented a review of cross-domain knowledge transfer methods for RL. In addition to the literature revision, works were categorized into a taxonomy based on their source of knowledge and knowledge transfer approach to facilitate the method-search process (see Table \ref{tab:taxonomy}). Additionally, works were characterized based on the data requirements and assumptions made to help find a method that fits with the resources available in a particular problem setting (see Table \ref{tab:comparison}).

From the data-requirement dimensions, the Inter-Domain Mapping dimension shows that over the years, methods have progressively required less human supervision. That is, older works that relied on an external entity providing the inter-domain state and/or action mapping were followed by methods that incorporated domain-agnostic alignment strategies (\textit{e.g.} manifold alignment \cite{wang2009manifold}). More recent works have developed methods to learn such mappings based on RL-specific elements such as the transition and reward dynamics. Among the most recent methods, there is an ongoing effort to make less restrictive assumptions (\textit{e.g.} not depending on paired demonstrations, autonomously selecting the best source of knowledge, learning from state-only demonstrations).

Given that more recent works are addressing settings with more relaxed data assumptions, we observe a growing interest in pushing the boundaries of autonomy in knowledge transfer. This shift towards unsupervised transfer will broadly impact areas such as robotics, where data scarcity is one of the main prevailing challenges that prevent learning robust policies for real-world environments \cite{chatzilygeroudis2017black,firoozi2023foundation}. As new robots are continuously introduced to the market, being able to transfer experiences across platforms will be a determining factor in the progress of robotics. Furthermore, although plenty of fundamental and practical challenges need to be addressed, adopting strategies that require less human intervention is a direction worth exploring, as self-management of resources, such as experiences, is a core ingredient of intelligence.

% Bibliography
\bibliographystyle{unsrt}
\bibliography{references}

\newpage

% Authors Biography
\begin{IEEEbiography}[{\includegraphics[width=1in,height=1.25in,clip,keepaspectratio]{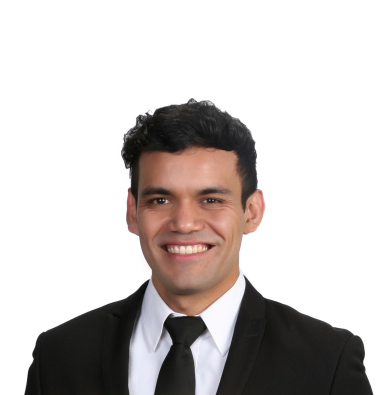}}]{Sergio A. Serrano} obtained his Bachelor's degree in Mechatronics Engineering in 2017 from the Instituto Tecnológico de Los Mochis, his Master's degree in Computer Science in 2019 from the Instituto Nacional de Astrofísica, Óptica y Electrónica (INAOE), and is currently studying his PhD in Computer Science from INAOE. He has collaborated in writing the latest edition of the \textit{Robótica de Servicio} (\textit{Service Robotics}) book and is an active member of the Service Robotics team from INAOE's robotics laboratory, which yearly participates at the \@Home category in the Mexican Robotics Tournament (TMR). His main research interest include robotics, reinforcement learning, transfer learning, computer vision and unsupervised learning.
\end{IEEEbiography}

\begin{IEEEbiography}[{\includegraphics[width=1in,height=1.25in,clip,keepaspectratio]{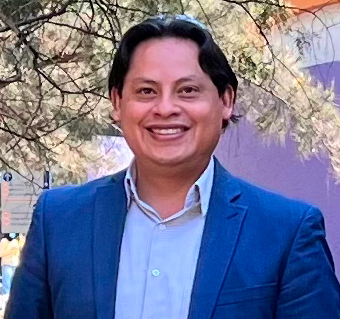}}]{Jose Martinez-Carranza} obtained his Bachelor's degree in Computer Science from the Benemérita Universidad Autónoma de Puebla, his Master's degree in Computer Science from the Instituto Nacional de Astrofísica, Óptica y Electrónica (INAOE), and his PhD from the University of Bristol in the United Kingdom. He currently holds the position of Researcher B at the Coordination of Computer Sciences at INAOE. In 2018, the University of Bristol awarded him the status of "Honorary Senior Research Fellow." In 2015, he received the Royal Society-Newton Advanced Fellowship distinction, awarded by the Newton Fund. He is the founder and leader of the Unmanned Aerial Systems research group, and his team, QuetzalC++, has participated in highly prestigious international competitions, winning various awards, notably the 1st Place in the IROS 2017 Autonomous Drone Racing competition and the 1st Place Regional (Latin America) in the OpenCV AI Competition 2021. He served as the General Chair of the 12th International Micro Air Vehicle Conference (IMAV) 2021, and also as an Associate Editor for the IEEE IROS international conference in 2022-2024. Since 2022, he has joined the editorial board of the scientific journal Unmanned Systems. His research includes aerial robotics and vision for robotics with topics such as visual SLAM, visual odometry and neural localisation.
\end{IEEEbiography}

\begin{IEEEbiography}[{\includegraphics[width=1in,height=1.25in,clip,keepaspectratio]{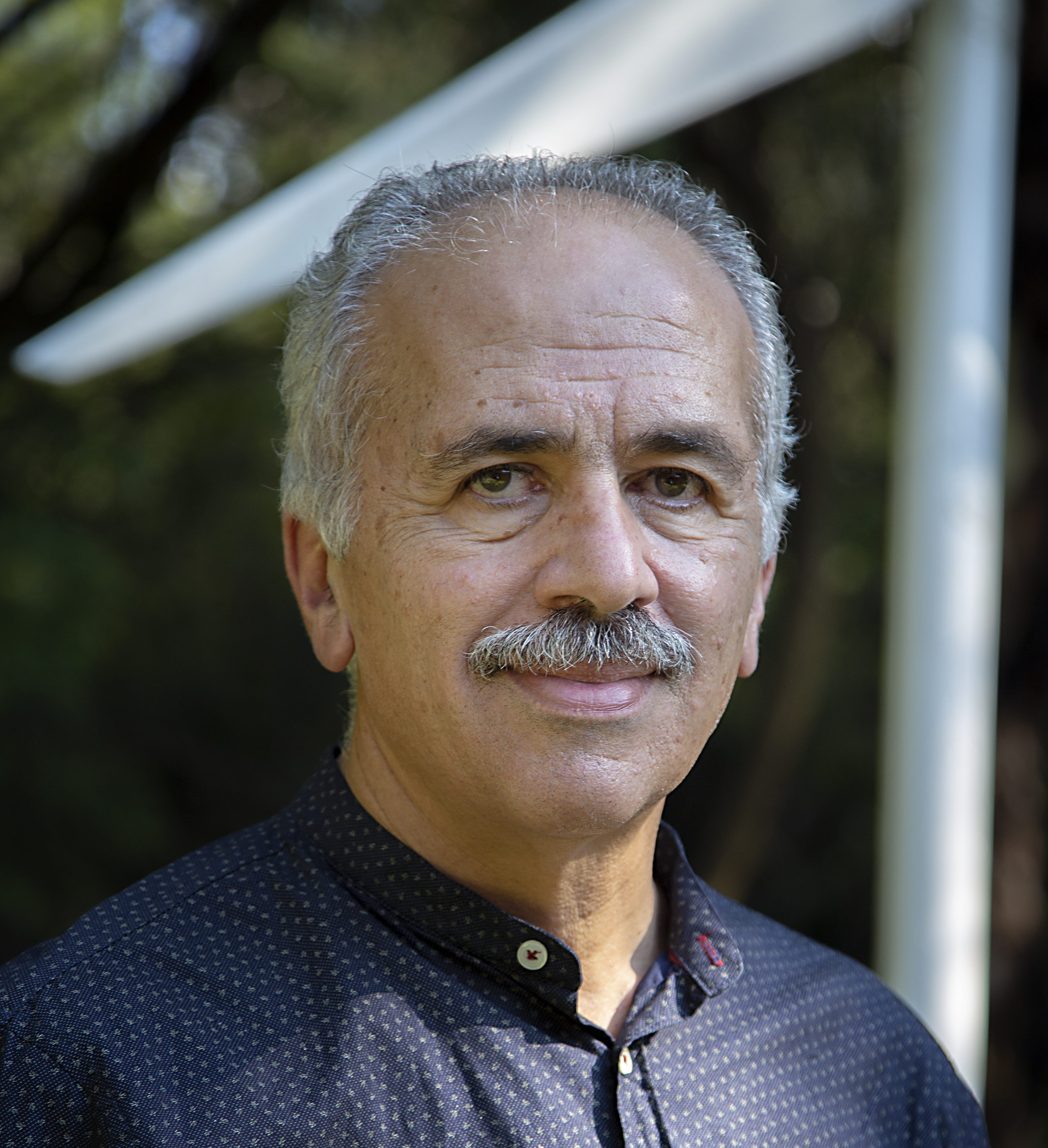}}]{L. Enrique Sucar} (Life Senior Member IEEE) has a Ph.D. in Computing from Imperial College, London, 1992; a M.Sc. in Electrical Engineering from Stanford University, USA, 1982; and a B.Sc. in Electronics and Communications Engineering from ITESM, Mexico, 1980. He has been a researcher at the Electrical Research Institute, a professor at ITESM, and is currently Senior Research Scientist at the National Institute for Astrophysics, Optics and Electronics, Puebla, Mexico. He has been an invited professor at the University of British Columbia, Canada; Imperial College, London; INRIA, France; and CREATE-NET, Italy. He has more than 400 publications and has directed nearly 100 Ph.D. and M.Sc. thesis. Dr. Sucar received the National Science Prize from the Mexican President; is Member Emeritus of the National Research System, and member of the Mexican Science Academy. He is associate editor of the Pattern Recognition, Computational Intelligence and Frontiers in Rehabilitation journals, and has served as president of the Mexican AI Society and the Mexican Academy of Computing. His main research interest are in probabilistic graphical models, causal reasoning and their applications in robotics, computer vision and biomedicine.
\end{IEEEbiography}

\end{document}